\documentclass[runningheads]{llncs}

\usepackage{eccv}

\usepackage{eccvabbrv}
\usepackage{graphicx}
\usepackage{booktabs}
\usepackage{wrapfig}
\usepackage{makecell}
\usepackage{subcaption}

\usepackage{amsmath,amsfonts,bm,amssymb}
\usepackage{algorithm}
\usepackage{algpseudocode}
\usepackage[table]{xcolor}
\usepackage{tcolorbox}
\usepackage{xcolor}
\usepackage{pifont}

\usepackage[accsupp]{axessibility}  

\usepackage{hyperref}

\usepackage{orcidlink}

\begin{document}

\title{R\textsuperscript{2}M: Real-Aware Residual Model Merging for Robust and Generalizable Deepfake Detection} 

\titlerunning{R\textsuperscript{2}M for Robust Deepfake Detection}

\author{Jinhee Park\inst{1,2}\textsuperscript{*}\orcidlink{0000-0001-6206-0832} \and
Guisik Kim\inst{2}\textsuperscript{*}\orcidlink{0000-0002-8254-0881} \and
Choongsang Cho\inst{2}\orcidlink{0000-0001-5952-5491} \and
Junseok Kwon\inst{1}\textsuperscript{\textdagger}\orcidlink{0000-0001-9526-7549}}

\authorrunning{J. Park, G. Kim, et al.}


\institute{
Department of Computer Science and Engineering, Chung-Ang University, Korea
\email{\{iv4084em,jskwon\}@cau.ac.kr}
\and
Korea Electronics Technology Institute (KETI), Korea \\
\email{specialre@naver.com}, \email{ideafisher@keti.re.kr}
}
\maketitle
\begingroup
\renewcommand\thefootnote{*}
\footnotetext{Equal contribution.\quad
\textsuperscript{\textdagger} Corresponding author: Junseok Kwon (jskwon@cau.ac.kr).}
\endgroup

\vspace{-4mm}
\begin{abstract}
Deepfake generators evolve rapidly, making exhaustive data collection and repeated retraining impractical. Unlike generic multi-task settings, deepfake specialists share a common binary objective (Real vs. Fake) and mainly differ in generator-specific artifacts. However, naive parameter arithmetic can induce unintended decision-boundary shifts, causing unstable ranking behavior and degraded AUC under domain shift.
We propose R\textsuperscript{2}M, a training-free merging framework that decomposes task vectors into a shared component and generator-specific residuals, linking parameter-space updates to logit-space behavior. Offline spectral construction identifies shared and residual subspaces, and online routing selects residuals via first-order gradient-residual alignment, predicting per-sample logit updates under linearization.
R\textsuperscript{2}M is both efficient and interpretable: expensive computations are performed offline, while online inference requires only a single forward–backward pass with lightweight inner-product routing. The same formulation enables diagnostic analysis through margin and routing statistics. Experiments demonstrate consistently strong performance across in-domain, cross-domain, and unseen settings, highlighting R\textsuperscript{2}M as an interpretable and scalable approach to training-free deepfake model merging.
\vspace{-5mm}
\end{abstract}

\vspace{-3mm}
\section{Introduction}
\vspace{-2mm}
Deepfakes have evolved from simple face splicing to photorealistic synthesis driven by modern generative models~\cite{li2019faceshifter, chen2020simswap, tolosana2020deepfakes, rombach2022high}. Recent systems preserve identity while enabling precise control over lip movements and facial expressions, and high-quality content can now be generated through simple prompts on widely accessible generative platforms and APIs~\cite{prajwal2020lip, park2025community}.
While these advances democratize content creation, they amplify the spread of harmful and deceptive media, including financial fraud, copyright infringement, and political disinformation. These risks underscore the urgent need for robust deepfake detection methods.
Because deepfake generation algorithms evolve rapidly, exhaustively collecting per-generator data and retraining a unified detector is fundamentally unsustainable. Even with sufficient data, jointly training on heterogeneous forgeries introduces gradient interference \cite{yu2020gradient, standley2020tasks}. Conversely, maintaining a dedicated specialist for each generator incurs substantial operational overhead \cite{shiohara2022detecting, yan2024df40}. 

To address these challenges, we adopt a \emph{model merging} paradigm for deepfake detection. Concretely, we first fine-tune specialist detectors on generator-specific data and subsequently merge them directly in parameter space to obtain a single unified model. This strategy enables rapid incorporation of new generators without revisiting previous training data or performing costly joint retraining \cite{izmailov2018averaging, wortsman2022model, ilharco2022editing}. To the best of our knowledge, this is the first systematic investigation of model merging for deepfake detection.

\begin{figure*}[t]
  \centering

 \begin{minipage}[t]{0.4\linewidth}  
    \centering
    \includegraphics[width=\linewidth]{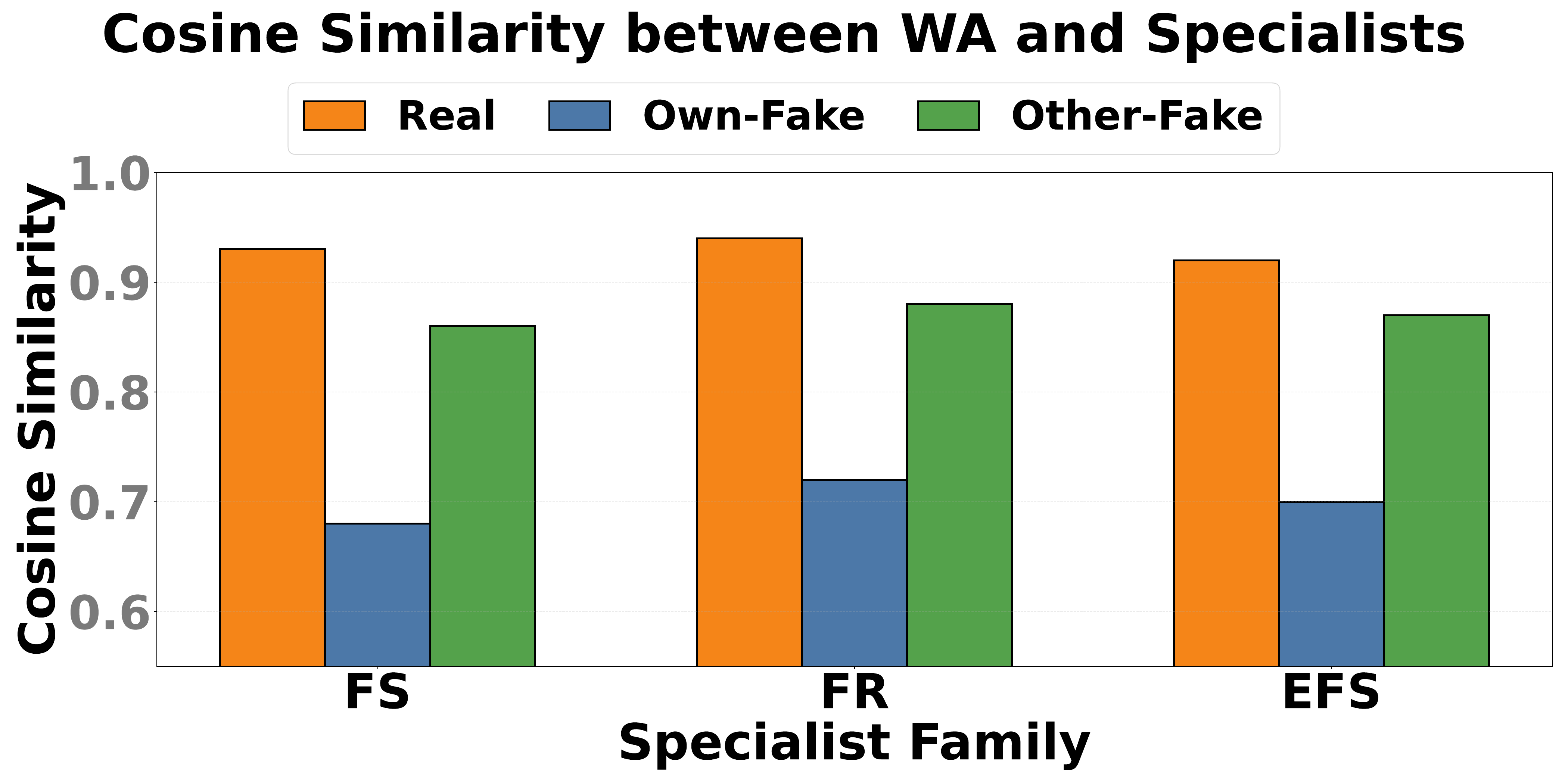} 
  \end{minipage}
    \hspace{3mm}
 \begin{minipage}[t]{0.53\linewidth}  
    \centering
    \includegraphics[width=\linewidth]{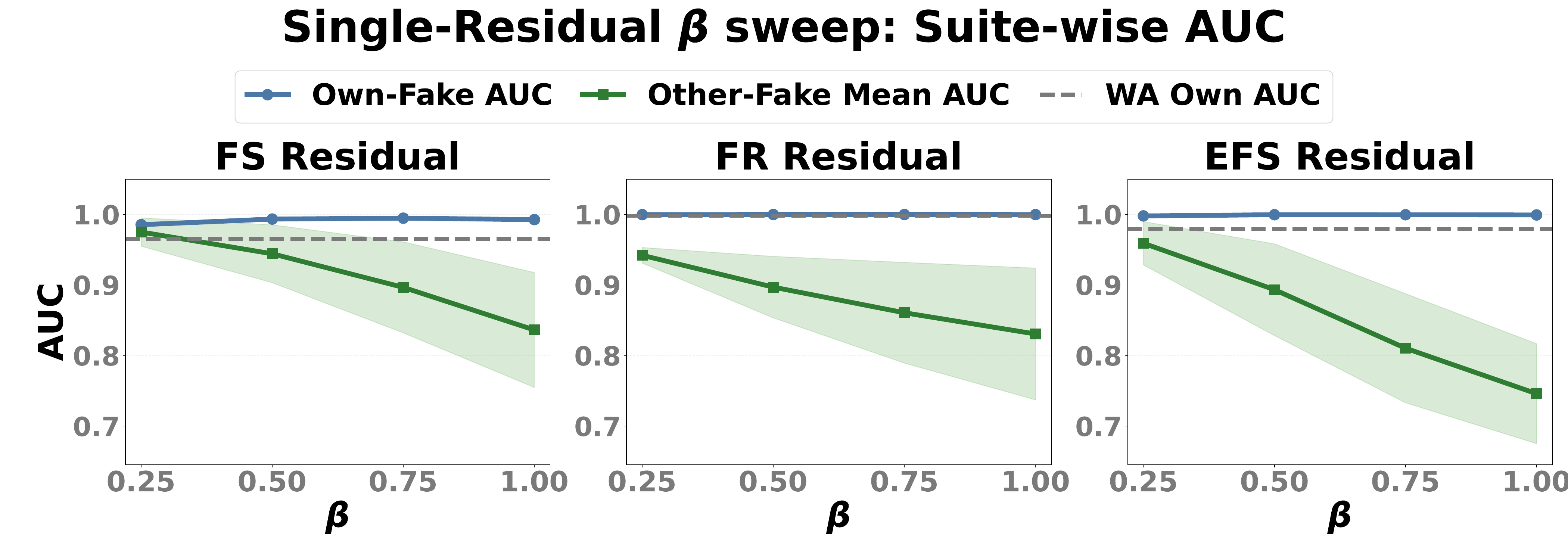} 
  \end{minipage}
  \begin{minipage}[t]{0.43\linewidth}
         \scriptsize {(a) Weight averaging preserves Real while suppressing Own-Fake alignment.}
  \end{minipage}
  \hspace{3mm}
  \begin{minipage}[t]{0.47\linewidth}   
         \scriptsize {(b) Residual gains are selective and conflicting.}
  \end{minipage}
      \vspace{-2mm}
  \caption{\textbf{Empirical observations motivating selective specialization in deepfake merging.}
  (a)  Weight averaging (WA) preserves shared Real structure while attenuating generator-specific fake cues. Cosine similarity between WA and each specialist in the pre-logit space remains high on \emph{Real}, drops on \emph{Own-Fake}, and is relatively higher on \emph{Other-Fake}.
 (b) Injecting a single specialist residual improves performance on its corresponding forgery family over WA (dashed), but degrades others, exposing an inherent trade-off under unconditional merging. Solid curves show AUC of \(\theta_{\mathrm{WA}} + \beta(\theta_i - \theta_{\mathrm{WA}})\) as \(\beta\) varies. Dashed horizontal lines indicate the WA baseline AUC on each dataset. The green shaded band shows the spread of AUC across non-target forgery families (from the lowest to the highest).}
  \label{fig:obs}
    \vspace{-7mm}
\end{figure*}

\textbf{Observation 1.}
To validate this design choice, we analyze how a weight-averaged (WA) model \cite{izmailov2018averaging} relates to its specialists. We construct three specialists corresponding to distinct forgery families in DF40 \cite{yan2024df40}—FaceSwap (FS), FaceReenactment (FR), and Entire Face Synthesis (EFS)—all finetuned from the same pretrained backbone. A WA model is obtained by averaging their parameters, without additional retraining. 
We measure feature similarity between specialists and the WA model across Real, Own-Fake(generated by the specialist’s corresponding family), and Other-Fake images (generated by the remaining families).
Interestingly, a consistent pattern emerges across all families. On Real images, specialists exhibit high similarity to the WA model, indicating that weight averaging preserves the shared Real representation. In contrast, similarity drops markedly on Own-Fake images, revealing that generator-specific signals—strongly encoded in individual specialists—are attenuated in the WA model. In contrast, similarity on Other-Fake images remains substantially higher than on Own-Fake images, as they lack the specialist’s dominant fake cues. As illustrated in \cref{fig:obs}(a), this behavior implies that WA preferentially retains the shared Real structure while suppressing generator-specific Fake residuals. This indicates that the components suppressed by WA encode generator-specific signals discarded during averaging.

\textbf{Observation 2.}
To understand the role of parameters discarded by weight averaging, let $\theta_i$ denote the parameters of specialist $i$ and $\theta_{\mathrm{WA}}$ the weight-averaged model. We define the residual as $r_i = \theta_i - \theta_{\mathrm{WA}}$ and construct interpolated models $\theta(\beta) = \theta_{\mathrm{WA}} + \beta r_i$, where $\beta \in \mathbb{R}$ controls the residual strength. We evaluate these models across the three forgery families.
As shown in \cref{fig:obs}(b), increasing $\beta$ consistently improves performance on the corresponding family relative to WA (dashed), while degrading performance on the others. Although $\beta=1$ recovers the original specialist, even moderate $\beta$ values exhibit pronounced trade-offs.
This behavior indicates that residuals encode family-specific gains that conflict under unconditional merging. Consequently, no single global parameter setting can simultaneously realize all residual benefits, suggesting that specialization should be applied selectively rather than uniformly.

\textbf{Proposed R$\textsuperscript{2}$M.}
Building on these observations, we tailor merging to the structural properties of deepfake detection. Unlike conventional multi-task settings with disjoint label spaces, all specialists here operate on the same binary label space (Real vs. Fake). In practice, Real cues are largely stable and shared across datasets, whereas Fake cues are generator-specific and heterogeneous. Observations 1 and 2 therefore suggest that Real-aligned directions should be preserved globally, while Fake-specific directions should be activated conditionally to prevent cross-family interference.
Accordingly, an appropriate merging strategy should:
(i) preserve the shared Real structure,
(ii) retain complementary generator-specific Fake knowledge, and
(iii) apply Fake-specific gains selectively based on the input.
We realize this via \textit{Real-aware Residual Model Merging (R\textsuperscript{2}M)}.
We first extract a shared Real-aligned component by applying low-rank factorization (\eg SVD) to the specialists’ task vectors, interpreting the dominant directions as a core Real subspace. Each specialist update is then decomposed into a Real-aligned component and a residual encoding generator-specific Fake evidence.
To enable efficient conditional specialization, we derive an input-conditional routing rule via a first-order margin approximation. As a result, R$\textsuperscript{2}$M selects residuals through a single-gradient routing mechanism, eliminating multiple forward/backward evaluations of specialist models while preserving generator-specific discrimination.


\textbf{Contributions.} Our contributions are summarized as follows: (1) We present the first systematic study of model merging for deepfake detection, identifying a structural asymmetry: Real representations are largely shared across specialists, whereas Fake cues are generator-specific and conflict under unconditional merging. (2) We provide a geometric interpretation linking parameter-space factorization to feature-space structure, explaining why weight averaging preserves Real alignment and how R\textsuperscript{2}M decouples Real and Fake effects. (3) We propose an efficient input-conditional routing mechanism derived from a first-order margin approximation, enabling selective residual application with a single backward pass.(4) Extensive experiments on DF40 and cross-dataset benchmarks demonstrate improved generalization and robustness to unseen forgery families.

\vspace{-5mm}
\section{Related works}
\vspace{-2mm}
\textbf{Deepfake detection.}
Deepfake technology has advanced rapidly over the past decade. Early methods focused on localized facial manipulations \cite{faceswap, li2020advancing, liu2023deepfacelab}, while GAN-based generators substantially improved realism \cite{choi2018stargan, thies2019deferred, richardson2021encoding}, followed by diffusion models that further enhanced fidelity and controllability \cite{rombach2022high}. Beyond face replacement, modern talking-head and reenactment systems preserve identity while enabling natural control over expressions, speech, and head motion \cite{nirkin2019fsgan, li2024talkinggaussian, mukhopadhyay2024diff2lip, guo2024liveportrait}. Commercial platforms such as Veo, Kling, and Wan have further lowered the barrier to high-quality synthetic video generation. While these advances support creative applications, they also introduce significant privacy and safety risks.
On the detection side, prior work spans blending-based augmentation \cite{li2020face, shiohara2022detecting, lin2024fake}, frequency-domain analysis \cite{jeong2022frepgan, tan2024frequency, zhou2024freqblender}, and generalization to \emph{unseen} forgeries \cite{yan2023ucf, choi2024exploiting, cui2025forensics}. Representative approaches include self-blending for pseudo-fake generation \cite{shiohara2022detecting}, shared forgery feature extraction \cite{yan2023ucf}, high-frequency modeling \cite{tan2024frequency}, and temporally aware curriculum learning \cite{lin2024fake}. Adaptor-based designs improve efficiency \cite{cui2025forensics}, and vision–language signals have also been explored \cite{sun2025towards}.
More recently, \cite{yan2024effort} employ an SVD-based decomposition within a vision foundation model to separate pretrained semantic structure from forgery-specific cues, freezing principal components while adapting the orthogonal residual subspace.


\noindent\textbf{Model merging.}
Model merging combines task-specific experts into a single model by operating directly in parameter space, typically without additional training. A basic strategy is weight averaging (WA) \cite{izmailov2018averaging}, which averages parameters across experts. In large-scale fine-tuning, WA underpins ``model soups'' and often improves accuracy and robustness without increasing inference cost \cite{wortsman2022model}.
Beyond uniform averaging, task arithmetic represents each task as a task vector—the difference between fine-tuned and pretrained weights—and modifies model behavior through vector addition or subtraction; combining multiple task vectors enables multi-task capability \cite{ilharco2022editing}. To mitigate interference, TIES-Merging trims small updates, resolves sign conflicts, and merges only sign-consistent parameters, improving multitask performance across modalities \cite{yadav2023ties}.
More recently, WA has been revisited from the task-vector perspective. \textsc{CART} centers task vectors around the weight average and applies low-rank approximation to reduce cross-task interference \cite{choi2024revisiting}. Orthogonal approaches such as \emph{Fisher-weighted averaging} incorporate local curvature information during merging \cite{matena2022merging}. Recent work also explores combining model merging with test-time adaptation, where task coefficients or model parameters are adjusted during inference using gradient-based objectives \cite{yang2023adamerging}. 

\noindent\textbf{Mixture-of-Experts routing.}
Mixture-of-Experts (MoE) models scale capacity by activating only a subset of expert modules per input using a learned router \cite{shazeer2017outrageously}, with later variants simplifying routing and improving training stability \cite{fedus2022switch}. 
Subsequent work improves routing stability and load balancing, including expert-choice routing variants \cite{zhou2022mixture}. 
Recent efforts also explore converting dense checkpoints into MoE-style models for efficient deployment \cite{zhu2024moe}.

\noindent\textbf{Positioning of Our Work.}
To the best of our knowledge, model merging has not been systematically studied for deepfake detection. Existing merging approaches are largely task-agnostic, treating task updates symmetrically. In contrast, deepfake detection exhibits a structural asymmetry: Real representations are largely shared across specialists, whereas Fake cues are generator-specific. R\textsuperscript{2}M exploits this property by preserving a shared Real component while selectively activating generator-specific residuals, yielding a unified detector that remains composable as new forgery families are introduced.

Our approach is complementary to training-time generalization and subspace factorization methods for AI-generated image detection \cite{yan2024effort}, but targets a different deployment setting: training-free merging of independently fine-tuned specialists. It is also conceptually related to Mixture-of-Experts routing and test-time adaptation, yet R\textsuperscript{2}M neither learns a router nor updates parameters at inference; gradients are used only as an analytic signal for input-conditional residual selection.

\vspace{-5mm}
\section{Method} \label{sec:method}
\vspace{-2mm}
Let $f_\theta(x) \in \mathbb{R}^2$ denote the logits of a binary deepfake detector 
parameterized by $\theta$ for an input image sample $x$, corresponding to the classes Real and Fake. We define the classification margin as
\vspace{-2mm}
\begin{equation}
    z(\theta, x) = f_\theta^{\text{fake}}(x) - f_\theta^{\text{real}}(x),
\vspace{-1mm}
\end{equation}
so that $z(\theta,x) > 0$ indicates Fake and $z(\theta,x) < 0$ indicates Real.
This margin formulation is natural for deepfake detection, which is typically evaluated using ranking-based metrics such as AUC.
Let $\theta_0$ denote a pretrained backbone parameter and $\theta_i = \theta_0 + \Delta_i$ denote a specialist parameter trained on forgery family $i$, where $\Delta_i$ is its task vector.
We decompose each task vector as $\Delta_i = c + r_i$, where $c$ represents a shared component and $r_i$ captures family-specific residual updates. This decomposition exposes the shared and family-specific structure of task vectors, making (P1) and (P2) analyzable through explicit geometric constraints.
We consider a merged model of the form
$\theta^*(x) = \theta_0 +  c + \beta r_{i(x)}$,
where $i(x)$ denotes an input-dependent routing rule selecting a residual component, and $\beta$ controls the global residual strength.
For convenience, we define the shared model:
$\theta_c := \theta_0 + c$.  
We then define the residual-induced margin shift as
\vspace{-2mm}
\begin{equation}\label{eq:margin_delta}
    \Delta z_i(x)=z(\theta_c + \beta r_i, x)-z(\theta_c, x),
\vspace{-1mm}
\end{equation}
which quantifies how the residual component $r_i$ perturbs the classification margin relative to the shared model.
Finally, let $\mathcal{D}$ denote the overall data distribution, 
which consists of Real and Fake samples. 
We denote by $\mathcal{D}_{\text{real}}$ the marginal distribution 
of Real samples, and by 
$\mathcal{D}_{\text{fake}}^{(i)}$ the conditional distribution 
of Fake samples from forgery family $i$.
Then, a desirable merged model should satisfy the following properties:

\noindent\textbf{(P1) Real invariance (residual robustness).}
$\mathbb{E}_{x\sim \mathcal{D}_{\text{real}}}\big[|\Delta z_i(x)|\big]\le \epsilon$,
for each residual $r_i$,
where $\epsilon > 0$ is a small tolerance controlling Real robustness.
This condition ensures that residual updates do not significantly perturb the margin on Real samples, preserving the shared Real decision structure and keeping residual effects fake-specific.

\noindent\textbf{(P2) Fake selective gain (own-family amplification).}
For Fake samples from family $i$,
$\mathbb{E}_{x\sim \mathcal{D}_{\text{fake}}^{(i)}}\big[\Delta z_i(x)\big]\ge m$,
and for $j \neq i$,
$\mathbb{E}_{x\sim \mathcal{D}_{\text{fake}}^{(i)}}\big[\Delta z_j(x)\big]\le m'$,
where $m>m' \ge 0$ are desired margin gains for own- and cross-family samples, respectively.
This enforces that each residual selectively increases the margin for its corresponding forgery family, while exerting limited influence on others.

\vspace{-2mm}
\begin{tcolorbox}[colback=gray!5,colframe=black,boxrule=0.3pt]
\vspace{-2mm}
\textbf{Design Objective.}
Our goal is to design a decomposition and routing mechanism that approximately satisfies (P1) and (P2).
\vspace{-2mm}
\end{tcolorbox}
\vspace{-3mm}
\vspace{-4mm}
\subsection{Real--Fake Structured Decomposition}
\vspace{-2mm}
We aim to construct a decomposition $\Delta_i = c + r_i$ that approximately satisfies (P1) real invariance and (P2) fake selective gain. We first characterize the underlying geometric conditions required for these properties, and then show that they naturally admit a spectral solution.

\noindent{\textbf{Optimal Residual Geometry under Real–Fake Constraints.}}
For the shared model $\theta_c = \theta_0 + c$, the residual-induced margin shift $\Delta z_i(x)$ is defined in \cref{eq:margin_delta}. Applying a first-order Taylor expansion, $z(\theta+\delta,x)\approx z(\theta,x)+\nabla_\theta z(\theta,x)^\top \delta$, we obtain
\vspace{-2mm}
\begin{equation}\label{eq:taylor_approx}
\Delta z_i(x)\approx\beta \,\nabla_\theta z(\theta_c,x)^\top r_i=\beta \langle g(x), r_i\rangle,
\vspace{-1mm}
\end{equation}
where $g(x):=\nabla_\theta z(\theta_c,x)$ denotes the gradient of the margin evaluated at the shared model.
Under this linearization, the effect of a residual direction $r_i$ is governed by its alignment with the gradient evaluated at the shared model $\theta_c$.
In other words, residual-induced margin shifts depend on how strongly $r_i$ aligns with the local geometry of the margin at $\theta_c$.

To analyze \textbf{(P1)} in a tractable manner, we consider the expected squared margin shift on Real samples. Minimizing the squared shift provides a smooth surrogate for controlling $|\Delta z_i(x)|$. Using the first-order approximation derived in \cref{eq:taylor_approx},
$\Delta z_i(x)^2\approx\beta^2 \langle g(x), r_i \rangle^2=\beta^2 r_i^\top g(x) g(x)^\top r_i$.
Taking expectation over Real samples yields
$\mathbb{E}_{x\sim\mathcal D_{\text{real}}}\left[\Delta z_i(x)^2\right]\approx\beta^2\,r_i^\top\Sigma_R\, r_i$,
where $\Sigma_R$ is the real gradient covariance matrix, \ie $\Sigma_R=\mathbb{E}_{x\sim\mathcal D_{\text{real}}}\left[g(x) g(x)^\top\right]$.
Thus, Real sensitivity is governed by the quadratic form induced by $\Sigma_R$. Let the eigendecomposition be $\Sigma_R = U_R \Lambda_R U_R^\top$. Directions associated with large eigenvalues correspond to parameter perturbations that strongly affect the Real margin.

To satisfy \textbf{(P1)} under the squared-shift surrogate, we require each residual to have limited Real sensitivity:
\vspace{-2mm}
\begin{equation}
r_i^\top \Sigma_R r_i \le \epsilon_R,\qquad \epsilon_R>0.
\label{eq:real_constraint}
\vspace{-1mm}
\end{equation}
A sufficient condition is to remove the dominant eigenspace of $\Sigma_R$, motivating the projection step introduced next.

For Fake samples from family $i$,
$\mathbb{E}_{x\sim\mathcal D_{\text{fake}}^{(i)}}\Delta z_i(x)\approx\beta \, r_i^\top\mathbb{E}_{x\sim\mathcal D_{\text{fake}}^{(i)}}[g(x)]$.
Let $\mu_{F,i}:=\mathbb{E}_{x\sim\mathcal D_{\text{fake}}^{(i)}}[g(x)].$
Then \textbf{(P2)} promotes residual directions that increase $r_i^\top \mu_{F,i}$, \ie directions aligned with the mean Fake gradient of family $i$.
Combining this with the Real-invariance constraint in~\cref{eq:real_constraint} yields the following constrained trade-off:
\vspace{-2mm}
\begin{equation}
\max_{r_i}
\quad
r_i^\top \mu_{F,i}
\quad
\text{s.t.}
\quad
r_i^\top \Sigma_R r_i \le \epsilon_R,
\label{eq:fake_real_tradeoff}
\vspace{-1mm}
\end{equation}
which formalizes the requirement that residual directions should align with fake-specific gradients while avoiding dominant real-sensitive modes.


\noindent{\textbf{Spectral Interpretation.}} The constrained optimization in \cref{eq:fake_real_tradeoff}
admits a closed-form solution characterized by the spectrum of $\Sigma_R$.
\vspace{-2mm}
\begin{proposition}[Optimal Residual under Real--Fake Trade-off]\label{prop:1}
Let $\Sigma_R = U \Lambda U^\top$ be the eigendecomposition of the real gradient covariance matrix, and let $\mu_{F,i}$ denote the fake-gradient mean of family $i$.
Consider the constrained optimization problem $\max_r \; r^\top \mu_{F,i} \quad \text{s.t.} \quad r^\top \Sigma_R r \le \epsilon_R$.
An optimal solution is given by
$r_i^\star\propto U \Lambda^\dagger U^\top \mu_{F,i}=\sum_{k=1}^d\frac{u_k^\top\mu_{F,i}}{\lambda_k}\, u_k$,
where $\lambda_k$ and $u_k$ denote the eigenvalues and eigenvectors of $\Sigma_R$, and $\Lambda^\dagger$ is the Moore--Penrose pseudoinverse of $\Lambda$.
\end{proposition}

\begin{tcolorbox}[colback=gray!5,colframe=black,boxrule=0.3pt]
\vspace{-2mm}
\textbf{Implication.}
The optimal residual scales each fake-gradient component by the inverse of its corresponding Real-sensitivity eigenvalue. Consequently, directions associated with large $\lambda_k$ (\ie dominant Real-sensitive modes) are suppressed, while alignment with fake-specific gradients is retained. The proof is provided in \ref{app:prop1_proof}.
\vspace{-2mm}
\end{tcolorbox}


While \cref{prop:1} characterizes the optimal direction for a single family $r_i$, practical merging requires a shared low-rank residual structure across families. We therefore seek a rank-$k$ subspace $W \in \mathbb{R}^{d \times k}$ such that each residual $r_i$ lies in $\mathrm{span}(W)$. A natural surrogate for jointly preserving fake-specific alignment is the aggregated projection energy:
\vspace{-2mm}
\begin{equation}\label{eq:prop2}
\sum_{i=1}^T \|W^\top \mu_{F,i}\|^2=\mathrm{Tr}(W^\top M_F W),\quad M_F := \sum_{i=1}^T\mu_{F,i}\mu_{F,i}^\top,
\vspace{-1mm}
\end{equation}
which measures how well the subspace $W$ captures the family-wise Fake-gradient means.
Using the trace formulation in~\cref{eq:prop2}, we next characterize the optimal low-rank subspace under the Real-sensitivity constraint.
\vspace{-2mm}
\begin{proposition}[Optimal Low-Rank Residual Subspace]\label{prop:2}
Let $\Sigma_R \succeq 0$ denote the real-gradient covariance matrix, and let $M_F$ be defined in~\cref{eq:prop2}. To obtain a shared rank-$k$ residual structure, we consider the constrained spectral problem:
$\max_{W^\top W = I_k}\quad\mathrm{Tr}(W^\top M_F W)\quad\text{s.t.}\quad\mathrm{Tr}(W^\top \Sigma_R W)\le \epsilon_R$.
Then, any optimal solution $W^\star$ spans the top-$k$ generalized eigenspace of $(M_F,\Sigma_R)$. Equivalently, $W^\star$ is obtained by spectral decomposition of the whitened matrix $\Sigma_R^{-1/2} M_F \Sigma_R^{-1/2}$.
\end{proposition}

\begin{tcolorbox}[colback=gray!5,colframe=black,boxrule=0.3pt]
\vspace{-2mm}
\textbf{Implication.}
The optimal residual subspace arises from a generalized spectral trade-off between fake-specific alignment and Real robustness. Whitening by $\Sigma_R^{-1/2}$ attenuates directions that strongly influence Real margins, while $M_F$ promotes directions shared across Fake families. Consequently, the learned subspace captures fake-discriminative variations that lie in the complement of dominant Real-sensitive modes. The proof is provided in \ref{app:prop2_proof}.
\vspace{-2mm}
\end{tcolorbox}

\vspace{-5mm}
\subsection{Spectrally-Motivated Practical Construction}
\label{sec:spectral_practical}
\vspace{-2mm}
Propositions~\ref{prop:1} and~\ref{prop:2} establish that the optimal residual structure is governed by a generalized spectral problem involving the Real-gradient covariance $\Sigma_R$ and the Fake second-moment matrix $M_F$. In particular, the optimal residual subspace $W^\star$ corresponds to the top-$k$ eigenspace of the whitened matrix $\Sigma_R^{-1/2} M_F \Sigma_R^{-1/2}$.
Directly computing $\Sigma_R$ and $M_F$ is infeasible in high-dimensional parameter space. We therefore construct an efficient low-rank approximation that preserves the same spectral geometry while remaining tractable.

Conceptually, Proposition~\ref{prop:2} reveals that our method performs PCA on Fake gradients after removing the geometry induced by Real gradients.
The whitening step \emph{(Stage 1)} suppresses Real-sensitive directions, and the subsequent spectral decomposition \emph{(Stage 2)} extracts dominant Fake-structured components in this normalized space. In practice, this corresponds to projecting out dominant Real modes and then performing spectral analysis on the residual task vectors.
Overall, the proposed two-stage procedure yields a truncated and sample-efficient approximation to the theoretically optimal generalized spectral solution.

\begin{figure}[t]
\centering
\includegraphics[width=0.9\linewidth]{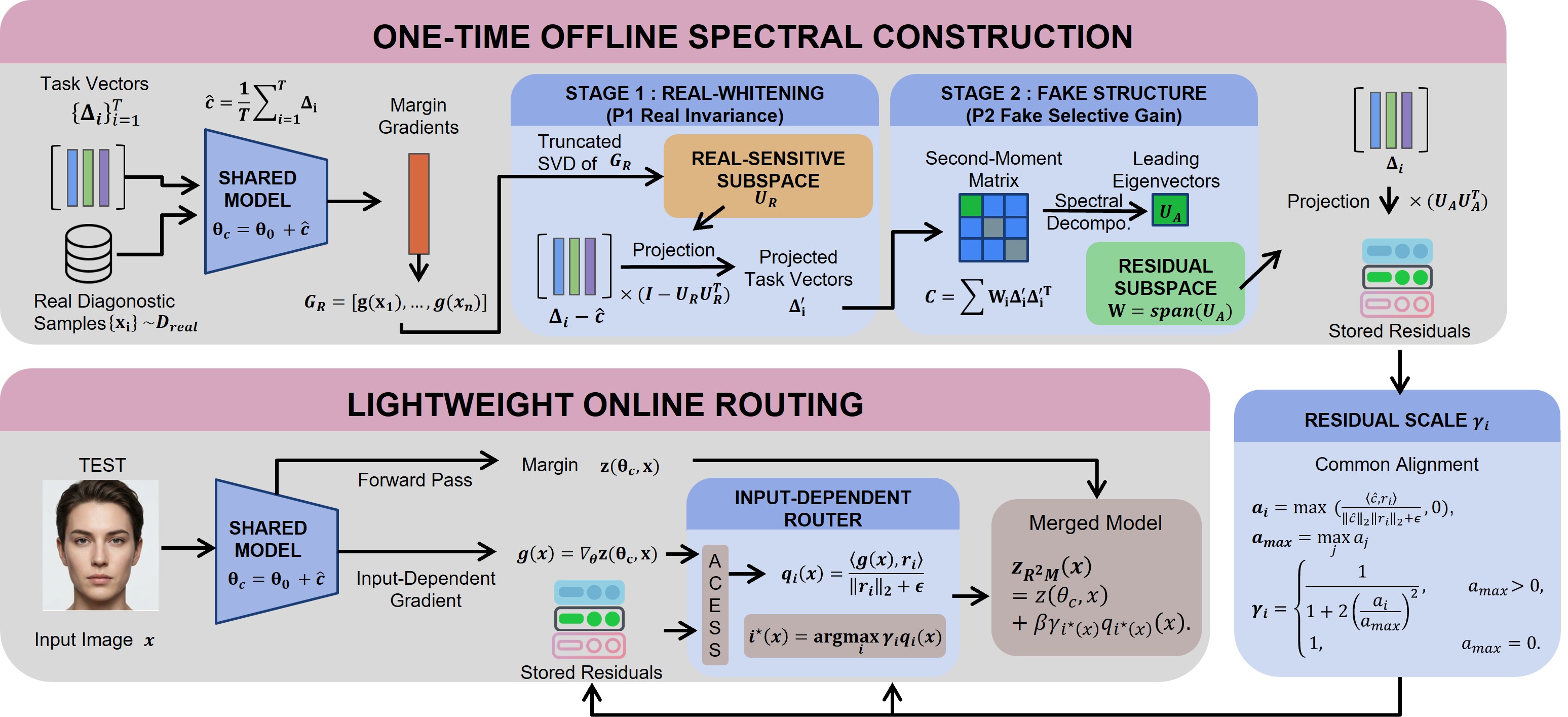}
\vspace{-3mm}
\caption{\textbf{Overall framework of the R\textsuperscript{2}M} 
The process is divided into two main phases: 
(Top) \textbf{One-time Offline Spectral Construction:} This phase identifies the optimal residual structure through two stages: 
\textit{Stage 1 (Approximate Real-Whitening)} suppresses dominant Real-sensitive modes to ensure P1; 
\textit{Stage 2 (Spectral Approximation)} extracts the leading Fake-structured components from projected task vectors to satisfy P2. 
(Bottom) \textbf{Lightweight Online Routing:} For a test sample $x$, a single backward pass obtains the margin gradient $g(x)$, allowing the input-dependent router to select the most aligned residual $r_{i^\star(x)}$ and approximate the merged margin $z_{\text{merged}}$.}
\label{fig:overall_framework}
\vspace{-7mm}
\end{figure}

\noindent{\textit{\textbf{Stage 1:} (Approximate Real-Whitening via Subspace Removal)}.}
Proposition~\ref{prop:1} indicates that optimal residual directions should attenuate the dominant eigendirections of $\Sigma_R$. Let $g(x)=\nabla_\theta z(\theta_c,x)$ denote the margin gradient evaluated at the shared model. Using a small set of Real diagnostic samples, we construct the gradient matrix
$G_R = [\, g(x_1), \dots, g(x_n) \,],\quad x_j \sim \mathcal D_{\text{real}}$.
Since $\Sigma_R \approx \frac{1}{n} G_R G_R^\top$, its principal eigenspace can be approximated via truncated SVD of $G_R$. Let $U_R$ denote the top-$r_0$ left singular vectors.  We first estimate the common component $\hat c=\frac{1}{T}\sum_{i=1}^T\Delta_i$ and then suppress real-sensitive modes by projecting each centered task vector onto the orthogonal complement:
$\Delta_i' = (I - U_R U_R^\top)(\Delta_i-\hat c)$.
This operation approximates whitening by suppressing 
the dominant eigenspace of $\Sigma_R$. 
Rather than explicitly rescaling by $1/\lambda_k$, 
we remove the principal Real-sensitive modes, 
which suffices to enforce (P1) in practice.

\noindent{\textit{\textbf{Stage 2:} (Spectral Approximation of Fake Second-Moment Structure)}.}
Proposition~\ref{prop:2} shows that the optimal residual subspace is determined by
the spectral structure of $M_F=\sum_i \mu_{F,i}\mu_{F,i}^\top$. In practice, the true Fake-gradient means $\mu_{F,i}$ are unknown and expensive to estimate precisely. Task vectors $\Delta_i$, being optimized to minimize family-specific Fake losses, naturally tend to align with $\mu_{F,i}$. We therefore use the projected task vectors $\Delta_i'$ as proxies for Fake-gradient directions.
To approximate $M_F$, we construct a weighted second-moment matrix in task space:
$C = \sum_{i=1}^T w_i\, \Delta_i' \Delta_i'^\top$,
where weights $w_i$ capture the degree of family-specific gradient alignment estimated from diagnostic samples. We then compute the top-$r_A$ eigenvectors of $C$, denoted by $U_A$. The final subspace $W = \mathrm{span}(U_A)$ 
serves as a low-rank surrogate for the leading eigenspace 
of $\Sigma_R^{-1/2} M_F \Sigma_R^{-1/2}$.

\vspace{-5mm}
\subsection{Practical Implementation}
\label{sec:practical_implementation}
\vspace{-2mm}
As summarized in \cref{fig:overall_framework,alg:r2m_practical}, R\textsuperscript{2}M consists of a \emph{one-time offline spectral construction} followed by \emph{lightweight online routing}. 
The offline phase estimates the common vector $\hat c$, removes Real-sensitive directions, extracts low-rank Fake-structured residuals, and stores a deterministic calibration scale for each residual branch. The online phase then uses a single margin-gradient computation to select one calibrated residual and update the common margin.

The calibration step accounts for approximation error in the practical construction. Since Real whitening is estimated from finite samples and Fake moments are replaced by task-vector proxies, a residual may still retain \emph{common-aligned leakage}. Such a branch can re-amplify shared evidence rather than contribute complementary Fake-specific evidence. We therefore shrink only residuals with positive alignment to the common vector. This correction is fixed after offline construction and does not introduce an additional learning objective.

\begin{algorithm}[t]
\caption{R\textsuperscript{2}M Offline Construction and Calibrated Routing}
\small
\textbf{Input:} pretrained parameter $\theta_0$, specialist task vectors $\{\Delta_i\}_{i=1}^T$, Real diagnostic samples, ranks $r_0,r_A$, residual strength $\beta$.\\
\textbf{Offline construction:}
\begin{enumerate}
\item Compute the common vector $\hat c=\frac{1}{T}\sum_{i=1}^T\Delta_i$ and shared model $\theta_c=\theta_0+\hat c$.
\item Estimate the Real-sensitive subspace $U_R$  from diagnostic gradients $G_R$.
\item Project centered task vectors away from Real-sensitive modes:\\
$
\Delta_i'=(I-U_RU_R^\top)(\Delta_i-\hat c).
$
\item Compute the top-$r_A$ eigenspace $U_A$ of
$
C=\sum_i w_i\Delta_i'\Delta_i'^\top
$
\\ and form residuals
$
r_i=U_AU_A^\top\Delta_i'.
$
\item Compute the positive common alignment
\\ $
a_i=\max\left(
\frac{\langle \hat c,r_i\rangle}{\|\hat c\|_2\|r_i\|_2+\epsilon},0
\right)
$
and $a_{\max}=\max_j a_j$.
\item Store residual scales
$
\gamma_i=
\begin{cases}
\frac{1}{1+2(a_i/a_{\max})^2}, & a_{\max}>0,\\
1, & a_{\max}=0.
\end{cases}
$
\end{enumerate}
\vspace{-2mm}
\textbf{Online routing:}
\vspace{-2mm}
\begin{enumerate}
\item For test input $x$, compute $z(\theta_c,x)$ and $g(x)=\nabla_\theta z(\theta_c,x)$.
\item Score each residual by
$
q_i(x)=\frac{\langle g(x),r_i\rangle}{\|r_i\|_2+\epsilon}.
$
\item Route to
$
i^\star(x)=\arg\max_i \gamma_i q_i(x)
$
and output
$
z_{\mathrm{R^2M}}(x)=z(\theta_c,x)+\beta\,\gamma_{i^\star(x)}q_{i^\star(x)}(x).
$
\end{enumerate}
\label{alg:r2m_practical}
\end{algorithm}

The scale $\gamma_i$ in \cref{alg:r2m_practical} acts as an inverse reliability penalty: the most common-aligned residual receives bounded shrinkage, while orthogonal or oppositional residuals remain unchanged. Since $\{\gamma_i\}$ are computed and stored offline, inference requires only one gradient computation and $T$ residual inner products per sample.

\vspace{-5mm}
\section{Experiments}\label{sec:exp}
\label{sec:exp-setup}
\vspace{-3mm}
\textbf{Benchmark and protocols (DF40).}
We evaluated on DF40~\cite{yan2024df40}, a recent deepfake detection benchmark that implements \emph{40} manipulation/generation methods across four \emph{forgery categories}: face swapping (FS), face reenactment (FR), entire-face synthesis (EFS), and face editing (FE). We followed DF40 naming of data ``domains'' (\eg FaceForensics++/FF~\cite{rossler2019faceforensics++} and Celeb-DF/CDF~\cite{li2020celeb}); we use FF and CDF as shorthand throughout the experiments. DF40 standardizes three evaluation protocols: (i) \textit{Protocol~1 (cross-forgery, same domain)}: train and test within the same data domain while varying forgery methods; (ii) \textit{Protocol~2 (cross-domain, same forgery)}: train and test on the same forgery category while changing the data domain; (iii) \textit{Protocol~3 (unknown forgery \& domain)}: train on seen forgeries/domains and test on \emph{unseen} forgeries and domains to simulate open-set conditions. These protocol definitions, the four-category taxonomy (FS/FR/EFS/FE), and FF/CDF domains were adopted from DF40.

\noindent\textbf{Specialist Model Construction and Training Policy.}
We trained three specialists, one per forgery category: $\theta_{\mathrm{FS}}$, $\theta_{\mathrm{FR}}$, and $\theta_{\mathrm{EFS}}$, each trained on the union of methods within its category (8 methods per category; 24 total). We also trained a single \emph{all-in-one} model on all 24 methods. All models use CLIP-L/14~\cite{radford2021learning} as the backbone with \texttt{pooler\_output} embeddings and a binary linear head trained via cross-entropy. Since our goal is to study training-free model merging for diverse forgery types, we train all specialists under an identical architecture and training recipe to ensure a controlled comparison. All training followed the official DF40 splits, with no data from Protocols 2 or 3 used for training or tuning. Training details and the full method list are provided in \ref{app:train-details}.

\noindent\textbf{Evaluation metrics.}
We reported image-level area under the ROC curve (AUC) under all three DF40 protocols and compared merged models against specialists and the all-in-one baseline. All evaluations followed the DF40 train/test splits and protocol definitions. We compared against training-free merging baselines.

\noindent\textbf{Model Merging Hyperparameter Selection.}
Hyperparameters were selected using Protocol~1 validation only and fixed across Protocols~2 and~3 to prevent cross-protocol leakage. For our method, the hyperparameters consist of two offline ranks and one online scaling factor. $r_0$ controls the dimension of the Real-sensitive subspace removed in Stage~1 (top-$r_0$ singular vectors $U_R$), $r_A$ determines the rank of the residual subspace extracted in Stage~2 (top-$r_A$ eigenvectors $U_A$ of $C$), and $\beta$ is the residual scaling coefficient applied at inference time. We searched $r_0 \in \left\{0,1,2\right\}$ and $r_A \in \left\{1,2,3\right\}$, and selected $\beta$ from $\left\{0, 6, 12, 18, 24, 30\right\}$. For baseline merging methods, we followed their hyperparameter grids; formulations are detailed in \ref{app:merging}. To ensure fair comparison, we applied the same search ranges across all merging methods with the same averaged head. Experiments related to the classifier head are included in \ref{app:supp-linear-head}.


\begin{table}[t]
\centering
\caption{\textbf{DF40 Protocols 1–2 Results (higher is better).} Columns are category$\times$domain $+$ domain-wise means. Protocol~1: cross-forgery within FF (in-domain).
Protocol~2: cross-domain transfer to Celeb-DF (CDF).
Gray cells denote category-matched specialist references, and blue cells denote the jointly trained all-in-one reference.
Bold and underline indicate the best and second-best results, respectively, computed only among training-free merging methods.}
\label{tab:df40-seen-auc}
\setlength{\tabcolsep}{7pt} 
\renewcommand{\arraystretch}{0.5}
\vspace{-3mm}
\scriptsize{
\begin{tabular}{lcccccccc}
\toprule
& \multicolumn{4}{c}{Protocol1 - FF} & \multicolumn{4}{c}{Protocol2 - CDF} \\
\cmidrule(lr){2-5}\cmidrule(lr){6-9}
Method & {FS} & {FR} & {EFS} & {\itshape Mean} & {FS} & {FR} & {EFS} & {\itshape Mean} \\
\midrule
\multicolumn{9}{l}{\emph{Our DF40-compliant training}}\\
Specialist--FS   & \cellcolor{gray!15}{0.995} & 0.912 & 0.766 & 0.891 &\cellcolor{gray!15}{0.959} & 0.690 & {0.603} & {0.751} \\
Specialist--FR   & 0.923 & \cellcolor{gray!15}{0.999} & 0.742 & 0.888 & 0.505 &\cellcolor{gray!15}{0.915} & 0.099 & 0.506 \\
Specialist--EFS  & 0.676 & 0.814 &\cellcolor{gray!15}{0.999} & 0.830 & 0.618 & 0.621 & \cellcolor{gray!15}{0.989} & {0.742} \\
\cellcolor{blue!8}All-in-one       & \cellcolor{blue!8}{0.962} & \cellcolor{blue!8}\cellcolor{blue!8}{0.997} & \cellcolor{blue!8}{0.978} & \cellcolor{blue!8}{0.979} & \cellcolor{blue!8}{0.759} & \cellcolor{blue!8}{0.860} & \cellcolor{blue!8}0.366 &\cellcolor{blue!8}0.662 \\
\midrule
\multicolumn{9}{l}{\emph{Training-free merging baselines}}\\
Weight Averaging   & 0.968 & \underline{0.997} & 0.982 & 0.982 & 0.825 & \underline{0.909} & 0.767 & 0.834 \\
Task Arithmetic    & 0.956 & 0.995 & 0.965 & 0.972 & 0.741 & 0.888 & \textbf{0.926} & 0.852 \\
Fisher        & 0.959 &  \textbf{0.998} & 0.949 & 0.969 & 0.735 & \textbf{0.919} & 0.442 & 0.699\\
CART               & \underline{0.976} & 0.994 & \underline{0.995} & \underline{0.988} & \underline{0.851} & 0.874 & \underline{0.907} & \underline{0.877} \\
\textbf{R$^{2}$M (ours)} &  \textbf{0.983} & 0.995 &  \textbf{0.998} &  \textbf{0.992} & \textbf{0.886} & \textbf{0.919} & 0.898 & \textbf{0.901} \\
\bottomrule
\end{tabular}
}
\vspace{-3mm}
\end{table}

\vspace{-5mm}
\subsection{Main Results on DF40 (Protocols 1–3)}
\label{subsec:main_exp}
\vspace{-3mm}
\cref{tab:df40-seen-auc} summarizes per-category AUCs under Protocols~1 and~2,
covering both in-domain (FF) and cross-domain (Celeb-DF) settings.

\begin{table}[t]
\centering
\caption{\textbf{DF40 Protocol 3 Results (higher is better).} \emph{Unseen datasets:} DeepFaceLab~\cite{liu2023deepfacelab}, HeyGen~\cite{heygen_site}, Midjourney~\cite{midjourney_home}, WhichIsReal~\cite{whichfaceisreal}, StarGAN~\cite{choi2018stargan}, StarGAN~v2~\cite{choi2020starganv2}, StyleCLIP~\cite{patashnik2021styleclip}, CollabDiff~\cite{huang2023collabdiff}.}
\label{tab:df40-unseen-auc}
\setlength{\tabcolsep}{4pt} 
\renewcommand{\arraystretch}{0.5}
\vspace{-3mm}
\scriptsize{
\begin{tabular}{lccccccccc}
\toprule
Method & \makecell{DeepFace\\Lab} & \makecell{Hey\\Gen} & \makecell{Mid\\journey} & \makecell{WhichIs\\Real} & \makecell{Star\\Gan} & \makecell{StarGan\\v2} & \makecell{Style\\Clip} & \makecell{Collab\\Diff} & \textit{Mean} \\
\midrule
\multicolumn{10}{l}{\emph{Our DF40-compliant training}}\\
Specialist--FS  & \cellcolor{gray!15}{0.892} & {0.573} & 0.334 & 0.364 & {0.887} & 0.640 & 0.608 & 0.596 & 0.612 \\
Specialist--FR  & 0.812 & \cellcolor{gray!15}0.546 & {0.603} & 0.228 & 0.697 & 0.433 & 0.053 & 0.141 & 0.439 \\
Specialist--EFS & 0.709 & 0.398 & \cellcolor{gray!15}{0.561} & \cellcolor{gray!15}{0.694} & \cellcolor{gray!15}{0.901} & \cellcolor{gray!15}{0.777} & \cellcolor{gray!15}{0.952} & \cellcolor{gray!15}{0.997} & {0.749} \\
\cellcolor{blue!8}All-in-one      & \cellcolor{blue!8}{0.884} & \cellcolor{blue!8}{0.561} & \cellcolor{blue!8}0.177 & \cellcolor{blue!8}{0.512} & \cellcolor{blue!8}0.860 & \cellcolor{blue!8}{0.714} & \cellcolor{blue!8}{0.761} & \cellcolor{blue!8}{0.854} & \cellcolor{blue!8}{0.665} \\
\midrule
\multicolumn{10}{l}{\emph{Training-free merging baselines (same backbone)}}\\
WA                       & \underline{0.930} & \textbf{0.626} & 0.613 & 0.455 & 0.953 & 0.728 & 0.635 & 0.848 & 0.724 \\
TA                       & 0.887 & 0.541 & \textbf{0.695} & 0.255 & 0.891 & 0.579 & 0.504 & 0.622 & 0.622 \\
Fisher                   & 0.896 & 0.623 &\underline{0.628} & 0.377 & 0.880 & 0.601 & 0.346 & 0.538 & 0.611 \\
CART                     & \textbf{0.943} & \underline{0.608} & 0.559 & \underline{0.497} & \textbf{0.975} & \underline{0.780} & \underline{0.813} & \underline{0.955} & \underline{0.766} \\
\textbf{R$^{2}$M (ours)} & 0.909 & 0.569 & 0.618 & \textbf{0.716} & \underline{0.958} & \textbf{0.817} & \textbf{0.962} & \textbf{0.997} & \textbf{0.818} \\
\bottomrule
\end{tabular}
}
\vspace{-7mm}
\end{table}

\noindent\textit{$\blacktriangleright$ Training-free model merging fits deepfake detection.}
Even plain \emph{Weight Averaging} achieves a strong FF mean AUC of \({0.982}\), essentially matching the jointly trained \emph{All-in-one} model (0.979).
This indicates substantial cross-task parameter sharing in this domain and suggests that weight-space merging is a natural and effective paradigm for deepfake detection.

\noindent\textit{$\blacktriangleright$ CART and R$^{2}$M retain specialists almost perfectly on FF (saturation).}
Both \emph{CART} and \emph{R$^{2}$M} achieve near-saturated performance on FF (Protocol~1), reaching mean AUCs of \({0.988}\) and \({0.992}\), respectively.
Category-wise values closely track the specialists (e.g., FS: 0.976/0.983 vs 0.995; FR: 0.994/0.995 vs 0.999; EFS: 0.995/0.998 vs 0.999).
This indicates that the proposed decomposition preserves specialist knowledge instead of washing out task-specific evidence through static averaging. Compared with training-free merging baselines, R\textsuperscript{2}M obtains the best mean AUC, showing that calibrated residual routing can recover category-specific gains without requiring additional training.

\noindent\textit{$\blacktriangleright$ Cross-domain seen transfer (CDF; Protocol~2).}
The advantage becomes more pronounced under Protocol~2 (Celeb-DF), where both the domain and forgery distribution differ from the FF training setting. In this regime, individual specialists and static merging baselines exhibit unstable transfer: For example, the FR specialist achieves an AUC of only \(0.099\) on EFS under CDF, indicating that decision boundaries learned for one manipulation category may not transfer across domain and manipulation shifts. Training-free merging methods mitigate such failures. In contrast, R\textsuperscript{2}M maintains a robust common component while selectively adding calibrated specialist residuals. As a result, it achieves the highest mean AUC  (\(0.901\)), suggesting that the proposed routing provides robustness under simultaneous domain and forgery shifts.

\cref{tab:df40-unseen-auc} reports per-forgery AUC on \emph{unseen} generators, with the macro mean.

\noindent\textit{$\blacktriangleright$ Merging prevents catastrophic failures and improves macro performance.}
The jointly trained \emph{All-in-one} model collapses on several unseen generators (e.g., \texttt{MidJourney} AUC \(=\) \(0.177\)). This suggests that a decision boundary learned for FR in FF does not transfer to EFS in Unseen domain,
indicating significant geometric misalignment across categories and domains.
Similarly, the jointly trained All-in-one model, while competitive on FF, drops to 0.366 on EFS under CDF, implying that collapsing heterogeneous forgeries into a single shared boundary can lead to suboptimal compromises under domain shift.
In contrast, all training-free merging variants avoid such failures.
For example, \emph{Weight Averaging} achieves \(0.613\) and \emph{CART} \(0.559\) on \texttt{MidJourney}.
This shows that merging preserves diverse specialist cues and improves robustness under distribution shift.

\noindent\textit{$\blacktriangleright$ R$^{2}$M delivers the strongest overall zero-shot transfer.}
Among training-free methods, \textbf{R$^{2}$M} achieves the best macro AUC on unseen generators (\(0.818\)), outperforming CART (\(0.766\)), Weight Averaging (\(0.724\)), and other baselines.
R$^{2}$M achieves the highest performance on multiple generators, including  \texttt{WhichIsReal} (\(0.716\)), \texttt{StarGAN v2} (\(0.817\)), \texttt{StyleCLIP} (\(0.962\)), and \texttt{CollabDiff} (\(0.997\)).
Overall, these results indicate that preserving manipulation-specific residual directions while controlling real-sensitive modes leads to stronger zero-shot generalization across diverse unseen generators.

\begin{table}[t]
\centering
\renewcommand{\arraystretch}{0.5}
\caption{\textbf{Effect of spectral ranks ($r_0$, $r_A$) and calibration.}
We report mean AUCs on sampled subsets of Protocol~1 (FF), Protocol~2 (CDF), and Protocol~3 (Unseen).
\textbf{Raw} uses unnormalized residual inner products, while \textbf{Cal} applies common-alignment calibration.
$\Delta$ denotes (Cal - Raw).}
\label{tab:rank_calibration}
\vspace{-3mm}
\scriptsize
\renewcommand{\arraystretch}{0.5}
\setlength{\tabcolsep}{3pt}
\begin{tabular}{cc|ccc|ccc|ccc}
\toprule
$r_0$ & $r_A$
& FF$_{\text{Raw}}$ & FF$_{\text{Cal}}$ & $\Delta$
& CDF$_{\text{Raw}}$ & CDF$_{\text{Cal}}$ & $\Delta$
& Uns$_{\text{Raw}}$ & Uns$_{\text{Cal}}$ & $\Delta$
\\
\midrule
0 & 1 & 0.898 & {0.990} & +0.092 & 0.398 & 0.830 & +0.433 & 0.696 & 0.792 & +0.096 \\
0 & 2 & 0.910 & 0.974 & +0.064 & 0.499 & 0.879 & +0.380 & 0.682 & 0.796 & +0.114  \\
0 & 3 & 0.913 & 0.975 & +0.062 & 0.511 & 0.880 & +0.369 & 0.684 & 0.798 & +0.114  \\
\midrule
1 & 1 & 0.944 & 0.988 & +0.044 & 0.436 & 0.851 & +0.415 & 0.705 & 0.787 & +0.082 \\
1 & 2 & 0.977 & 0.987 & +0.010 & 0.588 & {0.901} & +0.314 & 0.748 & 0.806 & +0.078 \\
1 & 3 & 0.984 & 0.989 & +0.005 & 0.589 & {0.901} & +0.312 & 0.749 & 0.812 & +0.063 \\
\midrule
2 & 1 & 0.944 & 0.988 & +0.044 & 0.436 & 0.851 & +0.416 & 0.708 & 0.789 & +0.081 \\
2 & 2 & 0.978 & 0.987 & +0.009 & 0.588 & 0.901 & +0.313 & 0.751 & {0.812} & +0.077 \\
\textbf{2} & \textbf{3}
& 0.980 & \textbf{0.992} & +0.012
& 0.601 & \textbf{0.901} & +0.300
& 0.755 & \textbf{0.818} & +0.063
\\
\bottomrule
\end{tabular}
\vspace{-7mm}
\end{table}

\vspace{-5mm}
\subsection{Ablation on Spectral Construction and Calibration}\label{subsec:ablations}
\label{subsec:ablations}
\vspace{-2mm}
 \cref{tab:rank_calibration} summarizes the mean AUCs across Protocols~1–3
for all $(r_0,r_A)$ configurations, before and after calibration. All results in this rank-grid analysis are obtained with a fixed residual scaling coefficient $\beta=18$. 
Therefore, performance differences arise solely from the spectral ranks $(r_0, r_A)$ and the presence or absence of calibration, not from tuning the residual magnitude.

\noindent\textbf{Effect of Spectral Ranks ($r_0$, $r_A$).}
Across both Raw and Cal settings, performance generally improves as the spectral ranks increase.
Using only one residual direction ($r_A=1$) is often insufficient, especially under domain shift, while increasing $r_A$ to $3$ yields stronger CDF and Unseen performance.
Similarly, increasing $r_0$ from $0$ to $2$ improves transfer performance, suggesting that removing Real-sensitive directions helps construct robust residual branches.
The gains saturate after moderate ranks, with $r_0=2$ and $r_A=3$ achieving the best results.

\noindent\textbf{Effect of Calibration.}
For every rank setting, Cal consistently outperforms Raw.
The improvement is modest on FF, where the test distribution is close to the specialist training domain, but becomes substantial on CDF, where Raw residual routing often drops below $0.60$.
This indicates that unnormalized residual inner products are sensitive to branch scale and common-aligned leakage under domain shift.
Residual normalization and common-alignment calibration mitigate this issue, producing more stable routing and stronger transfer performance.
Additional ablations on the residual scaling coefficient $\beta$ are provided in \ref{app:abl_beta}, showing that R\textsuperscript{2}M remains stable over a broad range of residual strengths and is not sensitive to the exact value of $\beta$.

\vspace{-3mm}
\subsection{Empirical Analysis of Real–Fake Decomposition}\label{subsec:assump_chk}
\vspace{-3mm}

\begin{figure}[t]
\centering
\includegraphics[width=\linewidth]{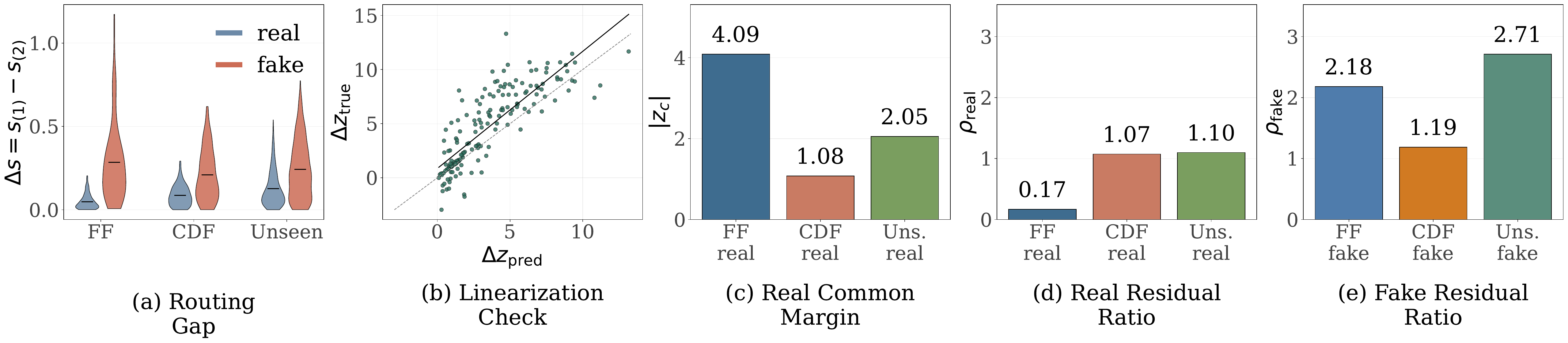}
\vspace{-7mm}
\caption{
Empirical analysis of the proposed real--fake decomposition.
(a) Calibrated routing confidence gap
$\Delta s=s_{(1)}-s_{(2)}$, where
$s_i(x)=\gamma_i\langle g(x),r_i\rangle/(\|r_i\|_2+\epsilon)$
and $s_{(1)}\ge s_{(2)}$ are the top-2 branch scores.
(b) Linearization check comparing the predicted residual shift
$\Delta z_{\mathrm{pred}}=\beta s_{i^\star}(x)$
with the actual logit shift $\Delta z_{\mathrm{true}}$ after applying the selected calibrated residual update.
(c) Common margin magnitude $|z_c|$ on real samples.
(d) Median residual-to-common ratio $\rho$ on real samples.
(e) Median residual-to-common ratio $\rho$ on fake samples.
}
\label{fig:decomposition}
\vspace{-7mm}
\end{figure}

\noindent\textbf{Routing selectivity.}
We first examine whether the residual branches provide selective evidence for individual inputs.
For each sample, we compute the calibrated branch score
$s_i(x)=\gamma_i\langle g(x),r_i\rangle/(\|r_i\|_2+\epsilon)$
and measure the top-2 confidence gap
$\Delta s=s_{(1)}-s_{(2)}$.
As shown in \cref{fig:decomposition}(a), fake samples have larger routing gaps than real samples across all domains, indicating more selective residual activation for manipulated inputs.
The separation is most pronounced on FF, where the real and fake distributions are well matched to the specialist training domains.
Under domain shift, the gap between real and fake becomes smaller on CDF and Unseen data, but fake samples still maintain higher routing confidence than real samples.
This suggests that residual branches retain manipulation-specific selectivity even beyond the source domain, although domain shift makes routing intrinsically more ambiguous.

\noindent\textbf{Linearization validity.}
\cref{fig:decomposition}(b) compares the predicted residual shift
$\Delta z_{\mathrm{pred}}=\beta s_{i^\star}(x)$
with the actual logit shift obtained by applying the selected residual update.
The two quantities are strongly correlated, showing that the gradient inner product is a reliable first-order surrogate for estimating the effect of residual updates.
This supports the use of gradient-based residual routing at test time.

\noindent\textbf{Common margin under real-domain shift.}
\cref{fig:decomposition}(c) reports the common margin magnitude $|z_c|$ on real samples.
The common model produces a large real margin on FF, but this margin decreases substantially on CDF and unseen domains.
This indicates that cross-domain real distribution shift weakens the shared common detector, making residual calibration more important.

\noindent\textbf{Residual-to-common ratio.}
\cref{fig:decomposition}(d,e) show the median residual-to-common ratio
$\rho=|\beta s_{i^\star}(x)|/(|z_c|+\epsilon)$
for real and fake samples, respectively.
On FF, real samples have a small residual ratio, while fake samples have a much larger ratio, suggesting that residual branches primarily contribute manipulation-specific evidence.
A similar separation is also observed on Unseen data, where fake samples maintain a substantially larger residual ratio than real samples.
This indicates that the residual directions capture fake-specific cues that transfer beyond the training forgery families.
On CDF, however, the gap between real and fake ratios becomes smaller because the common real margin collapses.
This explains why raw residual routing is unstable under cross-domain transfer: when real and fake residual ratios are less separated, branch scale mismatch and common-aligned leakage can more easily affect routing.
Common-alignment calibration mitigate this issue, leading to the large CDF gains observed in \cref{tab:rank_calibration}.

\vspace{-3mm}
\subsection{Merging Specialists with Heterogeneous Real Domains}
\label{subsec:heterogeneous_real_domains}
\vspace{-3mm}

\begin{wraptable}{r}{0.52\linewidth}
\vspace{-10mm}
\centering
\scriptsize
\setlength{\tabcolsep}{1pt}
\renewcommand{\arraystretch}{0.5}
\caption{\textbf{Merging six specialists with heterogeneous real distributions.}
AUC on FF/CDF domains for individual specialists and merged models.
Gray cells indicate the matched specialist reference.
R\textsuperscript{2}M-domain uses the same calibrated intra-group routing as R\textsuperscript{2}M after domain grouping.}
\label{tab:ff-cdf-merge}
\begin{tabular}{lcccccc}
\toprule
& \multicolumn{3}{c}{FF} & \multicolumn{3}{c}{CDF} \\
\cmidrule(lr){2-4}\cmidrule(lr){5-7}
Model & FS & FR & EFS & FS & FR & EFS \\
\midrule
\multicolumn{7}{l}{\emph{Individual specialists}}\\
FS-FF   & \cellcolor{gray!15}0.995 & 0.912 & 0.766 & 0.959 & 0.696 & 0.621 \\
FR-FF   & 0.923 & \cellcolor{gray!15}0.999 & 0.742 & 0.503 & 0.923 & 0.106 \\
EFS-FF  & 0.676 & 0.814 & \cellcolor{gray!15}0.999 & 0.635 & 0.608 & 0.988 \\
FS-CDF  & 0.754 & 0.647 & 0.616 & \cellcolor{gray!15}0.999 & 0.989 & 0.982 \\
FR-CDF  & 0.557 & 0.849 & 0.532 & 0.991 & \cellcolor{gray!15}0.999 & 0.929 \\
EFS-CDF & 0.329 & 0.244 & 0.923 & 0.955 & 0.961 & \cellcolor{gray!15}0.999 \\
\midrule
\multicolumn{7}{l}{\emph{Merged models}}\\
WA & 0.788 & 0.929 & 0.910 & 0.986 & 0.992 & 0.979 \\
CART & 0.867 & 0.957 & 0.954 & 0.998 & 0.994 & 0.996 \\
\cmidrule(lr){1-7}
\textbf{R\textsuperscript{2}M} unres. & 0.778 & 0.922 & 0.920 & {0.999} & 0.978 & 0.982 \\
\qquad domain & \textbf{0.929} & \textbf{0.988} & \textbf{0.979} & \textbf{0.999} & \textbf{0.997} & \textbf{0.997} \\
\bottomrule
\end{tabular}
\vspace{-6mm}
\end{wraptable}

In the main experiments above, the merged specialists differ by forgery type (FS, FR, EFS) but are trained under the same FF real domain.
Thus, they share a largely common real distribution, and R\textsuperscript{2}M only needs to route among forgery-specific residuals.
We further consider a harder stress test where this assumption is relaxed: we merge six specialists, FS/FR/EFS models trained from FF and FS/FR/EFS models trained from CDF.
Here, residual branches may carry both forgery-specific evidence and real-domain-specific response scale.
As shown in \cref{tab:ff-cdf-merge}, unrestricted routing over all six residuals preserves CDF performance but degrades FF performance.
Unlike a black-box merged detector, R\textsuperscript{2}M exposes the common vector, residual branches, and routing scores, allowing us to diagnose this failure directly.

\begin{figure}[t]
\centering
\includegraphics[width=0.9\linewidth]{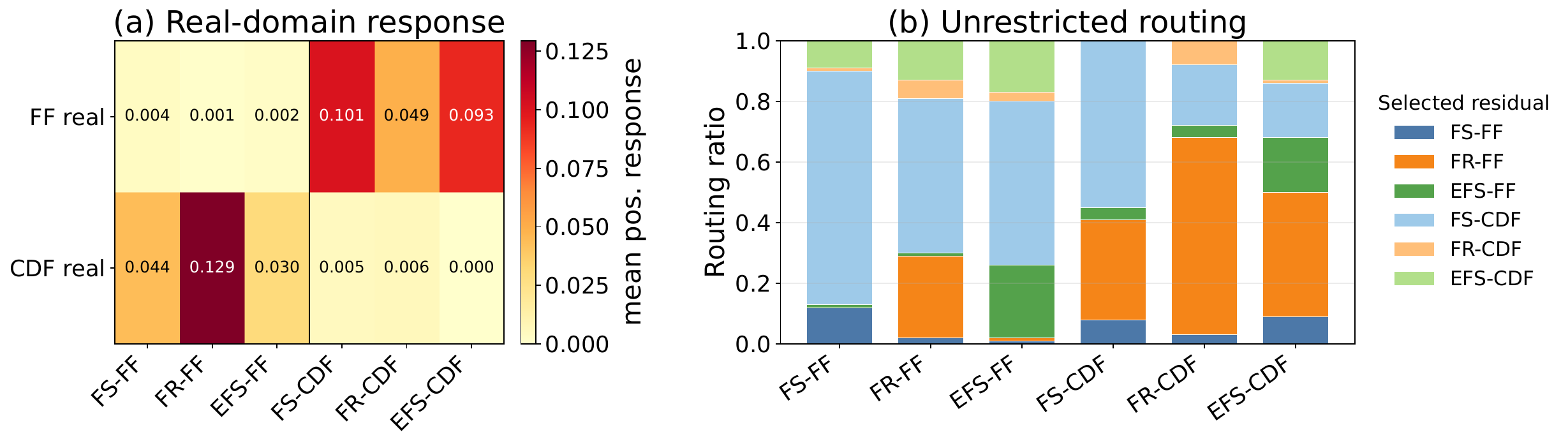}
\vspace{-4mm}
\caption{\textbf{Analysis of routing under heterogeneous real domains.}
(a) Mean positive residual response on real samples from each domain.
Opposite-domain residuals respond strongly, indicating domain-specific score-scale mismatch.
(b) Routing distribution without domain grouping.
FF inputs are often routed to CDF residuals, which explains the FF degradation in \cref{tab:ff-cdf-merge}.}
\label{fig:heterogeneous_diagnostic}
\vspace{-6mm}
\end{figure}

\cref{fig:heterogeneous_diagnostic} reveals the cause.
In \cref{fig:heterogeneous_diagnostic}(a), CDF-trained residuals activate strongly on FF real samples, while FF-trained residuals activate on CDF real samples.
\cref{fig:heterogeneous_diagnostic}(b) confirms that this response mismatch affects routing: under unrestricted routing, many FF inputs are assigned to CDF-trained residuals (\eg 77\% of FS-FF samples are routed to FS-CDF).
Thus, the degradation comes from domain-induced score-scale mismatch, not from a lack of useful residuals.

The diagnostic also gives a simple correction.
We compute a real-response contrast for each residual,
\(d_i=(\mathbb E_{\mathrm{CDF\ real}}[q_i]-\mathbb E_{\mathrm{FF\ real}}[q_i])/\mathrm{Std}_{\mathrm{real}}[q_i]\).
Residuals with \(d_i>0\) form the FF-trained group, while those with \(d_i<0\) form the CDF-trained group; no fake validation data is used.
After selecting the domain-consistent residual pool, we apply the original calibrated three-branch R\textsuperscript{2}M rule,
\(i^\star(x)=\arg\max_{i\in G}\gamma_i q_i(x)\).
This domain-grouped variant recovers FF performance and outperforms CART on all six sampled domains, showing that R\textsuperscript{2}M's interpretable decomposition can turn a failure mode into an effective correction.

\vspace{-4mm}
\section{Conclusion}
\vspace{-3mm}
We study model merging for deepfake detection, where specialists share a binary objective but capture generator-specific artifacts.
R\textsuperscript{2}M links parameter-space composition to logit-space behavior, showing that multiple specialist detectors can be consolidated without additional training data.
To the best of our knowledge, this is the first systematic study of this setting.
On DF40, R\textsuperscript{2}M achieves the best performance among training-free merging methods across in-domain, cross-domain, and unseen-generator evaluations.

Beyond accuracy, R\textsuperscript{2}M exposes an interpretable common--residual decomposition of the merged detector.
Its common vector, residuals, and routing scores allow us to diagnose failure modes, as shown in the heterogeneous-domain stress test, and to derive simple extensions such as domain-consistent residual grouping.
The main limitation is inference cost: input-level routing requires a gradient computation for each sample.
Developing faster routing mechanisms is therefore an important direction for future work.

\vspace{-4mm}
\section*{Acknowledgement}
\vspace{-3mm}
This work was supported by Institute of Information \& communications Technology Planning \& Evaluation (IITP) grant funded by the Korea government(MSIT) (No. RS-2024-00456709, No. RS-2025-02218768, No. RS-2022-II220926, No. RS-2026-25525363, No. RS-2021-II211341).

%
%
\bibliographystyle{splncs04}
\bibliography{main}

\newpage

\appendix
\setcounter{section}{0} 
\setcounter{figure}{0}
\renewcommand{\thefigure}{A\arabic{figure}}

\setcounter{table}{0}
\renewcommand{\thetable}{A\arabic{table}}

\setcounter{equation}{0}
\renewcommand{\theequation}{A\arabic{equation}}
\section{Theoretical and Algorithmic Details}

\subsection{Notation and setup}

Let $z(\theta,x)=f_\theta^{\mathrm{fake}}(x)-f_\theta^{\mathrm{real}}(x)$ denote the margin, and let the shared model be $\theta_c=\theta_0+ c$. Define the margin gradient at $\theta_c$ by
\begin{equation}
    g(x):=\nabla_\theta z(\theta_c,x)\in\mathbb{R}^d.
\end{equation}

For Fake samples from family $i$ and for Real samples, define
\begin{equation}
\mu_{F,i}:=\mathbb{E}_{x\sim\mathcal D_{\mathrm{fake}}^{(i)}}[g(x)],
\qquad
\Sigma_R:=\mathbb{E}_{x\sim\mathcal D_{\mathrm{real}}}\!\left[g(x)g(x)^\top\right],
\end{equation}
where $\Sigma_R\succeq 0$ by construction.
Under the first-order approximation
$\Delta z_i(x)\approx \beta\, g(x)^\top r$,
the Fake-selective/Real-invariant trade-off can be expressed as the constrained problem:
\begin{equation}
\max_{r\in\mathbb{R}^d}\quad r^\top \mu_{F,i}
\qquad
\text{s.t.}\quad
r^\top \Sigma_R r \le \epsilon_R,
\label{eq:app_tradeoff}
\end{equation}
where $\epsilon_R>0$ is a prescribed tolerance.

\subsection{Proof of Proposition 1}
\label{app:prop1_proof}

We consider the constrained optimization problem in \cref{eq:app_tradeoff}.
Since the objective is linear in $r$ and we consider directions within the effective (real-sensitive) subspace where the quadratic form is well-defined, the optimum is attained at the boundary, i.e., $r^\top \Sigma_R r=\epsilon_R$.
Form the Lagrangian
\begin{equation}
    \mathcal{L}(r,\lambda)=r^\top \mu_{F,i}-\lambda\,(r^\top \Sigma_R r-\epsilon_R),\qquad \lambda\ge 0.
\end{equation}
Stationarity with respect to $r$ yields
\begin{equation}
\nabla_r \mathcal{L}(r,\lambda)=\mu_{F,i}-2\lambda \Sigma_R r=0,\quad\Rightarrow\quad\mu_{F,i}=2\lambda \Sigma_R r.
\label{eq:app_stationarity}
\end{equation}
If $\Sigma_R$ is invertible,~\cref{eq:app_stationarity} gives
\(
r=\frac{1}{2\lambda}\Sigma_R^{-1}\mu_{F,i}.
\)
If $\Sigma_R$ is singular, a minimum-norm solution is given:
\begin{equation}
r=\frac{1}{2\lambda}\Sigma_R^{\dagger}\mu_{F,i},
\label{eq:app_pinv_solution}
\end{equation}
where $\Sigma_R^\dagger$ denotes the (Moore--Penrose) pseudoinverse of $\Sigma_R$. The multiplier $\lambda$ is uniquely determined by enforcing $r^\top \Sigma_R r=\epsilon_R$ (when the constraint is active), which fixes the scale of $r$ but does not change its direction.

Let $\Sigma_R = U\Lambda U^\top$ be an eigendecomposition, where $U=[u_1,\ldots,u_d]$ is orthonormal and
$\Lambda=\mathrm{diag}(\lambda_1,\ldots,\lambda_d)$ with $\lambda_1\ge\cdots\ge \lambda_d\ge 0$. For symmetric $\Sigma_R\succeq 0$, the Moore--Penrose pseudoinverse admits the spectral form $\Sigma_R^\dagger = U\Lambda^\dagger U^\top$, where $\Lambda^{\dagger}=\mathrm{diag}(\lambda_1^{\dagger},\ldots,\lambda_d^{\dagger})$ and $\lambda_k^{\dagger}=1/\lambda_k$ for $\lambda_k>0$ and $0$ otherwise.
Substituting into~\cref{eq:app_pinv_solution} yields
\begin{equation}
    r \propto U\Lambda^{\dagger}U^\top \mu_{F,i}.
\end{equation}
We have
\begin{equation}
r \propto U\Lambda^{\dagger}U^\top \mu_{F,i}= \sum_{j=1}^d \lambda_j^{\dagger}(u_j^\top \mu_{F,i}) u_j.
\end{equation}
For any eigenvector $u_k$, projecting onto $u_k$ and using orthonormality ($u_k^\top u_j=\delta_{kj}$, kronecker delta) yields
\begin{equation}
u_k^\top r\propto\lambda_k^{\dagger} (u_k^\top \mu_{F,i})=\frac{u_k^\top \mu_{F,i}}{\lambda_k}.
\end{equation}
\hfill $\square$

Propostion 1 thus establishes that the optimal residual direction is given by $r_i^\star \propto \Sigma_R^\dagger \mu_{F,i}$, which corresponds to inverse-spectral reweighting of the fake-gradient mean. As a consequence, directions that strongly affect Real margins (large $\lambda_k$) are automatically suppressed, while fake-specific alignment is preserved.

\subsection{Proof of Proposition 2}
\label{app:prop2_proof}

We formalize the subspace objective (Proposition2) introduced in the main text:
\begin{equation}\label{eq:app_prop2_primal}
\max_{W^\top W = I_k}
\quad
\sum_{i=1}^T \|W^\top \mu_{F,i}\|^2
\quad
\text{s.t.}
\quad
\mathrm{Tr}(W^\top \Sigma_R W)\le \epsilon_R.
\end{equation}

\begin{align}
\sum_{i=1}^T \|W^\top \mu_{F,i}\|^2
&=
\sum_{i=1}^T (W^\top \mu_{F,i})^\top (W^\top \mu_{F,i})\\
&=
\sum_{i=1}^T \mu_{F,i}^\top W W^\top \mu_{F,i}\\
&=
\sum_{i=1}^T \mathrm{Tr}(\mu_{F,i}^\top W W^\top \mu_{F,i})
\qquad \text{(scalar = trace)}\\
&=
\sum_{i=1}^T \mathrm{Tr}(W^\top \mu_{F,i}\mu_{F,i}^\top W)
\qquad \text{(cyclic trace)}\\
&=
\mathrm{Tr}\!\left(
W^\top
\Big(
\sum_{i=1}^T \mu_{F,i}\mu_{F,i}^\top
\Big)
W
\right)
\\
&=
\mathrm{Tr}(W^\top M_F W).
\end{align}
Assume the trace constraint in~\cref{eq:app_prop2_primal} is active at the optimum and that a corresponding Lagrange multiplier $\eta^\star \ge 0$ exists (KKT conditions hold). Since the objective is monotonically increasing in the subspace energy along fake-gradient directions, the trace constraint must be active at the optimum unless $M_F$ is orthogonal to the feasible set. Then the constrained problem is equivalent to maximizing the penalized objective
\begin{equation}
\max_{W^\top W = I_k}\ 
\mathrm{Tr}(W^\top M_F W) - \eta^\star \mathrm{Tr}(W^\top \Sigma_R W).
\end{equation}
Using linearity of trace, we can combine the two quadratic terms as
\begin{align}
\mathrm{Tr}(W^\top M_F W) - \eta^\star \mathrm{Tr}(W^\top \Sigma_R W)
&=
\mathrm{Tr}(W^\top M_F W) - \mathrm{Tr}(W^\top (\eta^\star \Sigma_R) W) \\
&=
\mathrm{Tr}\!\left(W^\top (M_F-\eta^\star \Sigma_R) W\right).
\end{align}
Let $A_{\eta^\star} := M_F - \eta^\star \Sigma_R$. The problem becomes
\begin{equation}
\max_{W^\top W = I_k}\ \mathrm{Tr}(W^\top A_{\eta^\star} W).
\label{eq:app_kfan}
\end{equation}
By the Ky Fan maximum principle, the maximum of~\cref{eq:app_kfan} equals the sum of the top-$k$ eigenvalues of $A_{\eta^\star}$, and any maximizer $W^\star$ can be chosen such that its columns are the corresponding top-$k$ eigenvectors of $A_{\eta^\star}$.

\paragraph{$\Sigma_R$-normalized formulation.}

For interpretational purposes, consider an equivalent $\Sigma_R$-normalized parameterization of the subspace.
Assume that the optimal subspace lies in a region where $\Sigma_R$ is positive definite (\eg after restricting to $\mathrm{range}(\Sigma_R)$). 
Then one can reparameterize its basis $W$ so that $W^\top \Sigma_R W = I_k$. Then any rank-$k$ subspace can be represented by a basis $W$ satisfying
\begin{equation}
W^\top \Sigma_R W = I_k.
\label{eq:sigma_normalized}
\end{equation}
Under this normalization, the objective becomes
\begin{equation}
\mathrm{Tr}\!\big(W^\top (M_F - \eta^\star \Sigma_R) W\big)
=
\mathrm{Tr}(W^\top M_F W)
-
\eta^\star \mathrm{Tr}(W^\top \Sigma_R W).
\end{equation}
Using~\cref{eq:sigma_normalized}, we obtain
\begin{equation}
\mathrm{Tr}\!\big(W^\top (M_F - \eta^\star \Sigma_R) W\big)
=
\mathrm{Tr}(W^\top M_F W)
-
\eta^\star k.
\label{eq:constant_shift}
\end{equation}
Since $-\eta^\star k$ is constant with respect to $W$, maximizing the penalized objective under the $\Sigma_R$-normalization~\cref{eq:sigma_normalized} is equivalent to maximizing
\begin{equation}
\max_{W^\top \Sigma_R W = I_k}
\ \mathrm{Tr}(W^\top M_F W).
\label{eq:normalized_form}
\end{equation}
Hence, both formulations induce the same optimal subspace.

\paragraph{Generalized eigenvalue characterization.}
We now characterize the stationary condition of the normalized problem~\eqref{eq:normalized_form}.
Form the Lagrangian with symmetric multiplier $\Lambda \in \mathbb{R}^{k\times k}$:
\begin{equation}
\mathcal{L}(W,\Lambda)
=
\mathrm{Tr}(W^\top M_F W)
-
\mathrm{Tr}\!\big(\Lambda(W^\top \Sigma_R W - I_k)\big).
\end{equation}
Taking derivative with respect to $W$ using the identity $\nabla_W \mathrm{Tr}(W^\top A W) = 2 A W$ for symmetric $A$, and setting to zero yields
\begin{equation}
M_F W = \Sigma_R W \Lambda.
\label{eq:gevp_matrix}
\end{equation}
Therefore, for each column $w$ of $W^\star$, there exists $\lambda$ such that
\begin{equation}
M_F w = \lambda\, \Sigma_R w,
\label{eq:gevp_vector}
\end{equation}
\ie $w$ is a generalized eigenvector of the matrix pair $(M_F,\Sigma_R)$.

\paragraph{Whitening interpretation.}
If $\Sigma_R \succ 0$  (or on its support), multiplying~\cref{eq:gevp_vector} by $\Sigma_R^{-1/2}$ gives
\begin{equation}
\Sigma_R^{-1/2} M_F \Sigma_R^{-1/2} \tilde{w}
=
\lambda \tilde{w},
\qquad
\tilde{w} := \Sigma_R^{1/2} w.
\end{equation}
Thus, we first compute the top-$k$ eigenvectors $\{\tilde w_j\}_{j=1}^k$ of the whitened matrix $\Sigma_R^{-1/2} M_F \Sigma_R^{-1/2}$, and map them back to the original coordinates via $w_j=\Sigma_R^{-1/2}\tilde w_j$. The optimal residual subspace is then $\mathrm{span}\{w_1,\dots,w_k\}$.
\hfill $\square$

\section{Additional Experimental Results}

\begin{figure*}[t]
    \centering
    \begin{subfigure}[t]{\linewidth}
        \centering
        \includegraphics[width=\linewidth]{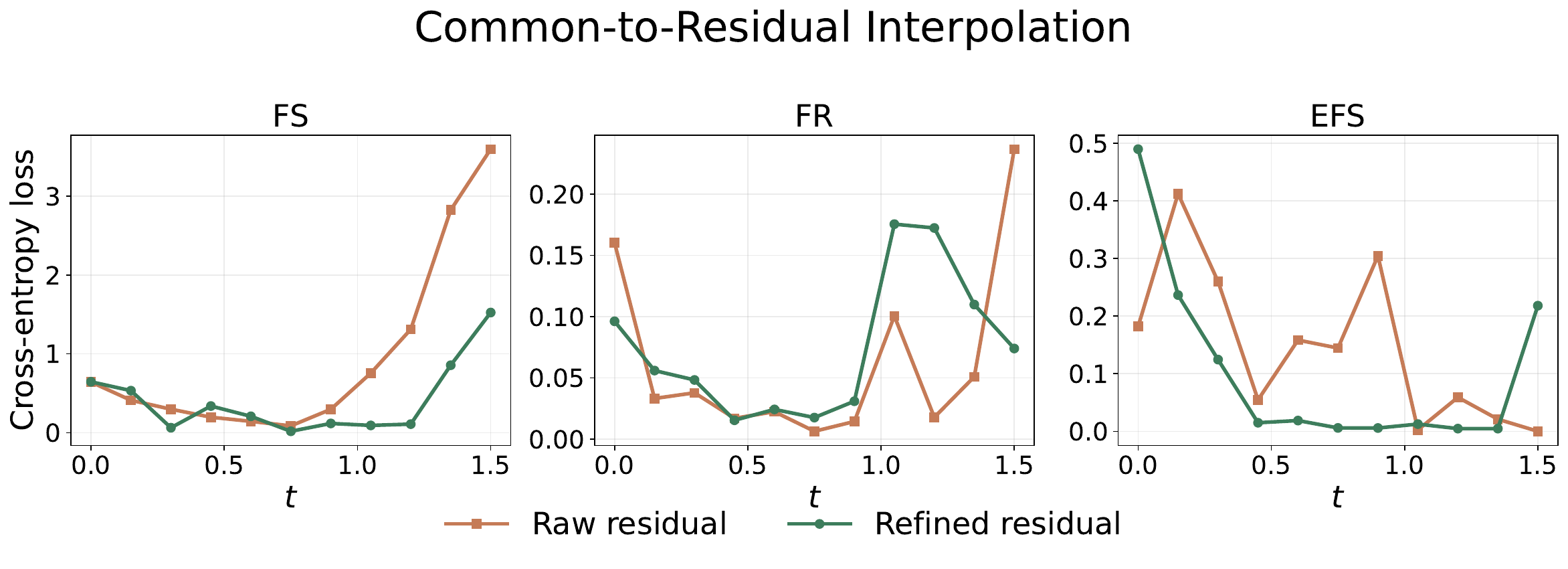}
        \caption{\textbf{Cross-entropy loss along the common-to-residual interpolation path.}}
        \label{fig:residual_interpolation_loss}
    \end{subfigure}

    \vspace{2mm}

    \begin{subfigure}[t]{\linewidth}
        \centering
        \includegraphics[width=\linewidth]{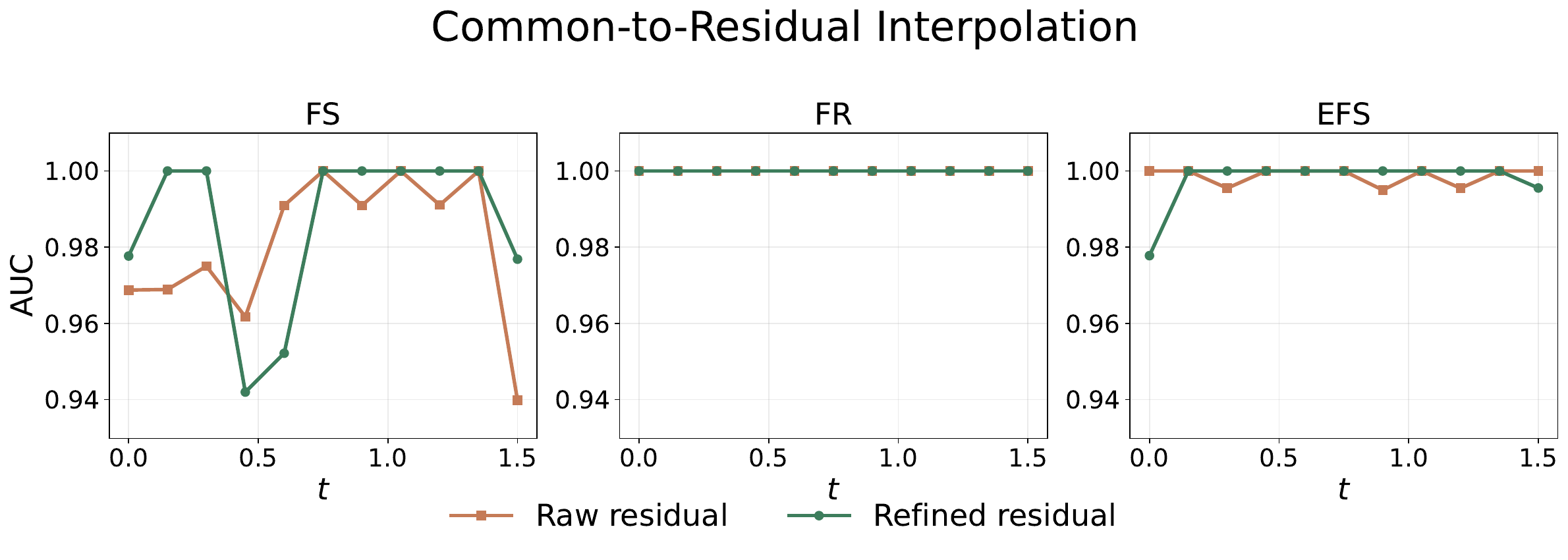}
        \caption{\textbf{AUC along the same common-to-residual interpolation path.}}
        \label{fig:residual_interpolation_auc}
    \end{subfigure}

    \caption{\textbf{Local interpolation from the common model along raw and refined residual directions.}
    For each FF specialist branch, we interpolate from the common model as $\theta(t)=\theta_c+t\,v$, where $v$ is either the raw centered residual $\Delta_i-c$ or the refined residual $U_AU_A^\top(I-P_{U_R})(\Delta_i-c)$.
    The top panel reports cross-entropy loss, and the bottom panel reports AUC along the same interpolation path.
    Overall, the refined residual yields a better-behaved local trajectory from the common model, with the clearest effect on the more difficult FS and EFS branches, while the difference is smaller for FR.}
    \label{fig:residual_interpolation}
\end{figure*}

\noindent\textbf{Local behavior of raw and refined residual branches.}
To directly examine the effect of residual refinement, we compare interpolation paths from the common model toward each specialist branch.
Specifically, for each specialist, we evaluate
$\theta(t)=\theta_c+t\,v$,
where $v$ is either the raw centered residual or the refined residual obtained after real-subspace removal and low-rank residual projection.
\Cref{fig:residual_interpolation} shows that the refined residual generally induces a better-behaved local trajectory from the common model.
This effect is most pronounced for the more difficult FS and EFS branches, where the refined direction yields a flatter loss profile while maintaining strong AUC.
By contrast, the difference is smaller for FR, whose common-to-specialist trajectory is already relatively stable even before refinement.

\noindent\textbf{Interpretation.}
These results are consistent with the role of our residual refinement procedure.
Instead of using the raw centered specialist difference directly, we remove real-sensitive components and retain only the dominant low-rank residual structure.
As a result, the resulting branch directions are better aligned with task-specific variation and induce less abrupt local degradation around the common model.
This supports the use of residual routing from the common anchor at inference time.

\noindent\textbf{Why residual refinement is necessary.}
A complementary observation comes from the raw residual routing results.
When the centered residuals are used directly without the proposed refinement, routing remains reasonably effective on FF, improving the mean AUC from 0.979 to 0.991.
However, the same strategy becomes highly unstable on CDF, where the mean AUC drops from 0.835 to 0.669.
This degradation is especially severe on EFS-CDF (0.735 $\rightarrow$ 0.277), showing that raw residual responses can become misaligned and overly dominant under domain shift.
These results further motivate our residual refinement procedure, which removes real-sensitive components and restricts routing to a more stable low-rank residual subspace.

\subsection{Training details} \label{app:train-details}

Let $\theta_0$ be the pretrained weights and $\tau_i=\theta_i-\theta_0$ the task vector of specialist $i$ among $N$.
\begin{itemize}
\item \textbf{Pretrained ($\theta_0$):} zero-shot without finetuning.
\item \textbf{Specialists ($\{\theta_i\}$):} one model per forgery method.
\item \textbf{All-in-one:} single model trained on the union of seen forgeries.

\end{itemize}

We strictly follow the official DF40 protocol without deviations in preprocessing, augmentation, optimization, or evaluation.

\paragraph{Protocol.}
\textbf{Video I/O:} H.264 compression level \texttt{c23}; we sample \textbf{8 frames per clip} for both training and testing at \textbf{224$\times$224} resolution. 
\textbf{Batching and system:} batch size 16 for training and testing; 1 GPU; \texttt{manualSeed=1024}. 
\textbf{Inputs:} no masks (\texttt{with\_mask=false}) and no facial landmarks (\texttt{with\_landmark=false}). 
\textbf{Normalization:} per-channel mean and standard deviation are set to (0.5, 0.5, 0.5).

\paragraph{Data augmentation.}
Random horizontal flip ($p{=}0.5$); small rotations within $\pm10^\circ$ ($p{=}0.5$); Gaussian blur with kernel size 3–7 ($p{=}0.5$); brightness and contrast jitter within $\pm0.1$ (each with $p{=}0.5$); JPEG quality jitter in [40, 100].

\paragraph{Optimization.}
Adam by default with \texttt{lr=$1\times10^{-5}$}, $\beta_1{=}0.9$, $\beta_2{=}0.999$, $\epsilon{=}10^{-8}$, and weight decay $5{\times}10^{-4}$. 
When SGD is used, we set \texttt{lr=$2\times10^{-4}$}, momentum 0.9, and the same weight decay. 
We do not use a learning-rate scheduler. Training runs for \textbf{3 epochs}.

\paragraph{Specialist task sets.} \hfill\\
\textbf{Task = 3} (three specialists; 24 total):

\emph{FS (8):} \texttt{fsgan}, \texttt{faceswap}, \texttt{facedancer}, \texttt{blendface}, \texttt{simswap}, \texttt{mobileswap}, \texttt{e4s}, \texttt{inswap}. 

\emph{FR (8):} \texttt{MRAA}, \texttt{facevid2vid}, \texttt{fomm}, \texttt{sadtalker}, \texttt{hyperreenact}, \texttt{mcnet}, \texttt{one\_shot\_free}, \texttt{wav2lip}.

\emph{EFS (8):} \texttt{SiT}, \texttt{ddpm}, \texttt{DiT}, \texttt{sd2.1}, \texttt{pixart}, \texttt{rddm}, \texttt{VQGAN}, \texttt{StyleGAN2}.
\hfill\\
\textbf{Task = 6} (six specialists; 31 total):

\emph{S1}: FS (5) = \texttt{uniface}, \texttt{simswap}, \texttt{mobileswap}, \texttt{faceswap}, \texttt{fsgan}. 

\emph{S2}: FS (4) = \texttt{inswap}, \texttt{blendface}, \texttt{e4s}, \texttt{facedancer}. 

\emph{S3}: FR (6) = \texttt{sadtalker}, \texttt{tpsm}, \texttt{fomm}, \texttt{MRAA}, \texttt{facevid2vid}, \texttt{pirender}. 

\emph{S4}: FR (6) = \texttt{hyperreenact}, \texttt{danet}, \texttt{lia}, \texttt{mcnet}, \texttt{one\_shot\_free}, \texttt{wav2lip}. 

\emph{S5}: EFS (5) = \texttt{pixart}, \texttt{StyleGANXL}, \texttt{StyleGAN3}, \texttt{DiT}, \texttt{ddpm}. 

\emph{S6}: EFS (5) = \texttt{rddm}, \texttt{StyleGAN2}, \texttt{SiT}, \texttt{VQGAN}, \texttt{sd2.1}.

\paragraph{Details of the heterogeneous Real experiment (Table 4)}\label{sec:supp-hetero-real}

\textbf{Datasets and Real/Fake composition.}

We construct three further specialists whose Real data come from CDF (FS-cdf, FR-cdf, EFS-cdf).
Since DF40 does not use CDF for training in its original protocol, we randomly split all available CDF videos into train/test/val sets for both Real and Fake, yielding:
\begin{itemize}
  \item FSAll\_Real:  1,282 / 320 / 1,602; \quad FSAll\_Fake: 4,129 / 1,032 / 5,161.
  \item FRAll\_Real: 1,709 / 427 / 2,136; \quad FRAll\_Fake: 6,216 / 1,554 / 7,770.
  \item EFSAll\_Real: 1,424 / 356 / 1,780; \quad EFSAll\_Fake: 7,104 / 1,776 / 8,880.
\end{itemize}
Each CDF-based specialist (FS-cdf, FR-cdf, EFS-cdf) is trained on the corresponding split using exactly the same architecture, data augmentations, and optimization hyperparameters as the FF based specialists.

\paragraph{Datasets and citations.}

\texttt{fsgan}~\cite{nirkin2019fsgan},
\texttt{faceswap}~\cite{faceswap},
\texttt{simswap}~\cite{chen2020simswap},
\texttt{facedancer}~\cite{rosberg2023facedancer},
\texttt{blendface}~\cite{shiohara2023blendface},
\texttt{mobileswap}~\cite{xu2022mobilefaceswap},
\texttt{e4s}~\cite{liu2023e4s},
\texttt{inswap}~\cite{inswapper}.
\texttt{uniface}~\cite{xu2022uniface},
\texttt{pirender}~\cite{ren2021pirenderer},
\texttt{danet}~\cite{hong2022dagan},
\texttt{lia}~\cite{wang2022lia},
\texttt{tpsm}~\cite{zhao2022tpsm},
\texttt{MRAA}~\cite{siarohin2021mraa},
\texttt{facevid2vid}~\cite{wang2019fewshotvid2vid},
\texttt{fomm}~\cite{siarohin2019fomm},
\texttt{sadtalker}~\cite{zhang2023sadtalker},
\texttt{hyperreenact}~\cite{bounareli2023hyperreenact},
\texttt{mcnet}~\cite{hong2023mcnet},
\texttt{one\_shot\_free}~\cite{wang2021oneshotfreeview},
\texttt{wav2lip}~\cite{prajwal2020lip}.
\texttt{VQGAN}~\cite{esser2021taming},
\texttt{StyleGAN2}~\cite{karras2020stylegan2},
\texttt{StyleGAN3}~\cite{karras2021stylegan3},
\texttt{StyleGANXL}~\cite{sauer2022styleganxl},
\texttt{sd2.1}~\cite{rombach2022ldm},
\texttt{ddpm}~\cite{ho2020ddpm},
\texttt{rddm}~\cite{liu2023rddm},
\texttt{pixart}~\cite{chen2024pixart},
\texttt{DiT}~\cite{peebles2023dit},
\texttt{SiT}~\cite{atito2021sit}.

\subsection{Model Merging Baselines}\label{app:merging}
\textbf{Model merging baselines (closed-form; no retraining).}
For every merged backbone $\hat{\theta}$, we attach the same averaged specialist head,
$\bar{\phi}=\tfrac{1}{N}\sum_{i=1}^{N}\phi_i$,
and evaluate the resulting model $(\hat{\theta},\bar{\phi})$ for all merging variants.
We consider the following baselines:

\begin{itemize}
    \item \emph{Weight Averaging (WA):}
    \[
    \theta_{\mathrm{avg}}=\theta_0+\frac{1}{N}\sum_{i=1}^{N}\tau_i,
    \]
    with no additional hyperparameters.

    \item \emph{Task Arithmetic (TA):}
    \[
    \theta_{\mathrm{ta}}(\alpha)=\theta_0+\alpha\cdot\frac{1}{N}\sum_{i=1}^{N}\tau_i,
    \]
    where $\alpha\in\{0.5,1.0\}$.

    \item \emph{Fisher Merging:}
    We use diagonal Fisher information to weight each specialist parameter coordinate.
    Let $\theta_i=\theta_0+\tau_i$ denote the $i$-th specialist backbone, and let
    $F_i$ be the diagonal Fisher estimate computed on the corresponding task data.
    The merged parameter is
    \[
    \theta_{\mathrm{fisher}}
    =
    \frac{\sum_{i=1}^{N} F_i \odot \theta_i}
    {\sum_{i=1}^{N} F_i+\epsilon},
    \]
    where $\odot$ denotes element-wise multiplication.
    Following standard Fisher merging, we estimate $F_i$ by the average squared gradient of the task loss and normalize each Fisher diagonal by its mean before merging.

    \item \emph{CART (origin-shifted low-rank merging):}
    Starting from $\theta_{\mathrm{avg}}$, we shift the origin, apply layer-wise SVD truncation with rank ratio $r$, and scale the resulting low-rank update:
    \[
    \theta_{\mathrm{cart}}(\eta,r)=\theta_{\mathrm{avg}}+\eta\,\hat{\tau}^{(r)},
    \]
    where $\eta\in\{0.5,1.0\}$ and $r\in\{0.1,0.3,0.5,0.7\}$.
\end{itemize}

For each baseline, we follow the original paper and official repository, while adapting the final evaluation interface to our unified protocol: for any merged backbone, we attach the same averaged specialist head $\bar{\phi}=\tfrac{1}{N}\sum_{i=1}^{N}\phi_i$.
  
\subsection{Ablation on beta}\label{app:abl_beta}

\begin{table}[h]
\centering
\caption{\textbf{Effect of residual scaling factor $\beta$ under $r_0=2$ and $r_A=3$.}
We report mean AUCs on sampled subsets of Protocol~1 (FF), Protocol~2 (CDF), and Protocol~3 (Unseen).
Raw uses unnormalized residual routing, while Cal applies residual normalization and common-alignment calibration.
All denotes the mean over FF/CDF/Unseen.}
\label{tab:beta_calibration}
\scriptsize
\renewcommand{\arraystretch}{0.5}
\setlength{\tabcolsep}{3pt}
\begin{tabular}{c|cc|cc|cc|cc}
\toprule
$\beta$
& FF$_{\text{Raw}}$ & FF$_{\text{Cal}}$
& CDF$_{\text{Raw}}$ & CDF$_{\text{Cal}}$
& Uns$_{\text{Raw}}$ & Uns$_{\text{Cal}}$\\
\midrule
0  & 0.985 & 0.985 & 0.820 & 0.820 & 0.749 & 0.749 \\
6  & 0.982 & {0.991} & 0.606 & 0.878 & 0.762 & 0.817  \\
12 & 0.980 & 0.990 & 0.592 & 0.896 & 0.754 & 0.817  \\
18 & 0.978 & \textbf{0.992} & 0.588 &  \textbf{0.901} & 0.751 & \textbf{0.818}  \\
24 & 0.978 & 0.985 & 0.586 & {0.900} & 0.749 & 0.815 \\
30 & 0.977 & 0.982 & 0.584 & {0.900} & 0.748 & 0.809  \\
\bottomrule
\end{tabular}
\end{table}

\begin{table}[h]
\centering
\caption{\textbf{Effect of classifier head choice.}
We evaluate R\textsuperscript{2}M with $r_0=2$, $r_A=3$, and $\beta=18$ while changing only the binary classifier head.
FF, CDF, and Unseen denote mean AUCs on sampled subsets of Protocols~1--3.
All denotes the mean of FF/CDF/Unseen.}
\label{tab:head_ablation}
\scriptsize
\renewcommand{\arraystretch}{0.5}
\setlength{\tabcolsep}{5pt}
\begin{tabular}{lccc}
\toprule
Head & FF & CDF & Unseen  \\
\midrule
Averaged head & \textbf{0.992} & \textbf{0.901} & \textbf{0.818} \\
FS head       & 0.987 & \textbf{0.901} & \textbf{0.818}  \\
FR head       & 0.987 & 0.899 & 0.815  \\
EFS head      & 0.988 & 0.900 & 0.817  \\
\bottomrule
\end{tabular}
\end{table}
\noindent\textbf{Effect of $\beta$.}
\Cref{tab:beta_calibration} evaluates the residual scaling factor $\beta$
under the fixed spectral setting $r_0=2$ and $r_A=3$.
When $\beta=0$, R\textsuperscript{2}M reduces to the common model without residual contribution.
As $\beta$ increases, calibrated routing substantially improves CDF and Unseen performance while preserving high FF accuracy.
In contrast, Raw routing becomes increasingly unstable under domain shift, especially on CDF, because uncalibrated residual scores are dominated by branch scale.
The calibrated results are stable over a broad range of $\beta$ values, with similar overall performance for $\beta\in[12,24]$.
We therefore use $\beta=18$ as a fixed default in the main experiments.
\subsection{Effect of different linear heads after merging} \label{app:supp-linear-head}

\noindent\textbf{Effect of head choice.}
\Cref{tab:head_ablation} compares the averaged classifier head with specialist-specific heads under the same R\textsuperscript{2}M routing setup.
All head choices produce nearly identical performance, with less than \(0.005\) difference in the overall mean AUC.
The averaged head is competitive with all specialist heads and avoids selecting a task-specific classifier after merging.
This suggests that the main gain of R\textsuperscript{2}M comes from common--residual decomposition and calibrated routing, rather than from classifier-head specialization.

\begin{figure}[t]
    \centering
    \includegraphics[width=\linewidth]{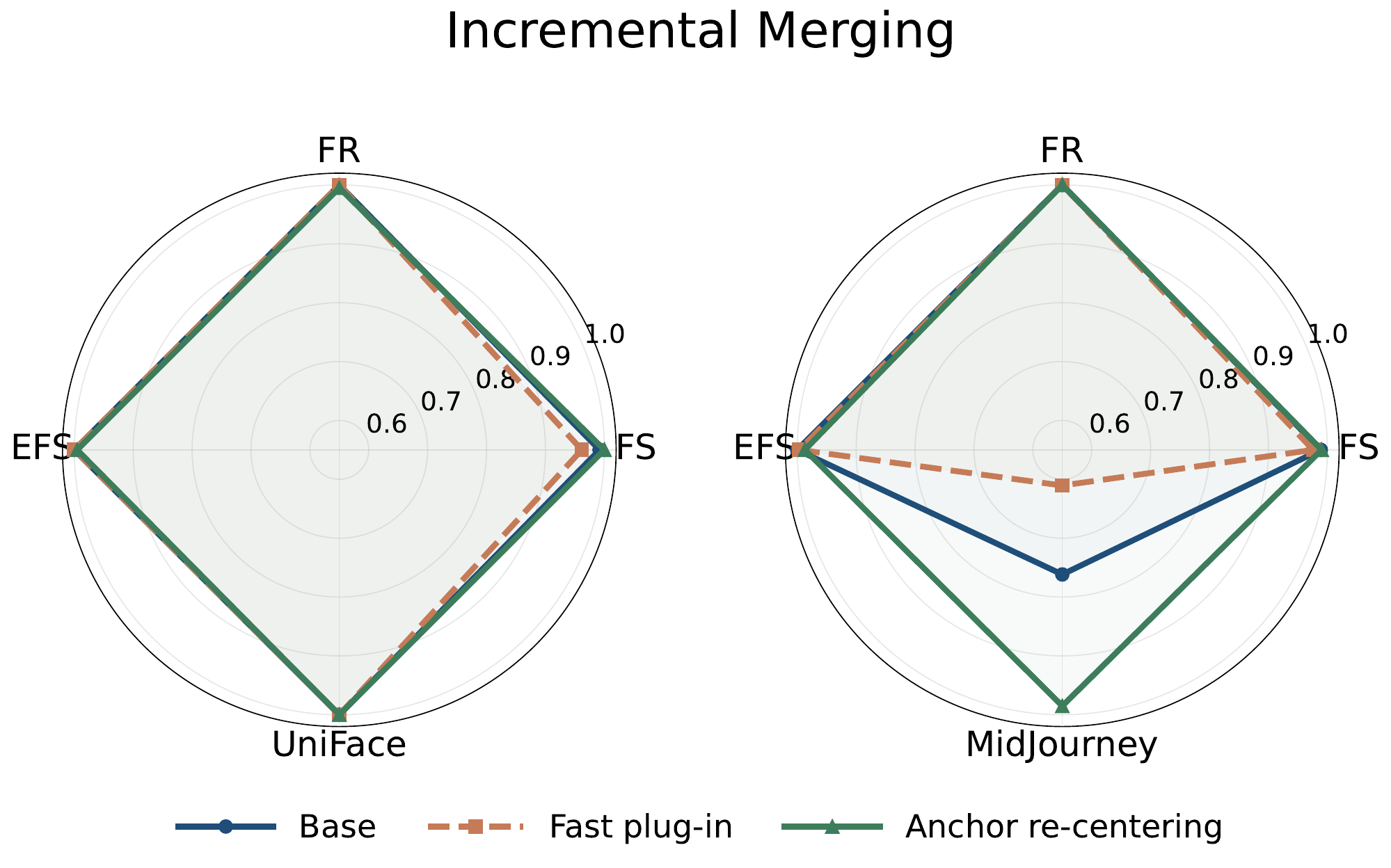}
\caption{\textbf{Incremental merging under nearby and novel emerging families.}
We compare three update modes: the original three-specialist R\textsuperscript{2}M system (\emph{Base}), adding only a new residual branch while keeping the original anchor fixed (\emph{Fast plug-in}), and rerunning the offline decomposition over the enlarged specialist pool (\emph{Anchor re-centering}).
Each radar plot reports AUC on the original FF families (FS, FR, EFS) and the newly introduced family.
For the nearby UniFace family, the base system already provides near-perfect coverage, so both update modes bring little additional gain.
For the more novel MidJourney family, fast plug-in is insufficient, whereas anchor re-centering substantially improves the new-family AUC while preserving performance on the original FF families.}
    \label{fig:incremental_emerging}
\end{figure}

\subsection{Incremental Merging}
\label{app:scaling}

We further consider an incremental-merging scenario where a new fake family becomes available after the initial merged detector has already been constructed.
The base R\textsuperscript{2}M system is built from the original FF specialists (FS, FR, and EFS) with $r_0=2$, $r_A=3$, and $\beta=18$.
We refer to the shared model $\theta_c=\theta_0+\hat c$ as the \emph{anchor}, since all residual branches are constructed around this point.
At inference time, we use the same common-alignment calibrated routing rule as in the main experiments.

We study two emerging families, UniFace and MidJourney, which are not used when constructing the initial FF-based system.
UniFace is a nearby family that is visually and structurally closer to the original FF manipulation types, while MidJourney is a more novel family that is less aligned with the initial residual geometry.

For each family, we compare three expansion modes.
\textbf{Base} directly evaluates the original three-specialist R\textsuperscript{2}M system on the new unseen family, without adding any new branch.
\textbf{Fast plug-in} keeps the original anchor $\theta_c$, Real-sensitive subspace, and residual subspace fixed, and constructs only a new residual branch for the additional specialist with respect to the original anchor.
\textbf{Anchor re-centering} reruns the offline R\textsuperscript{2}M construction over the enlarged specialist pool, recomputing the common vector, anchor, and residual branches.
Thus, fast plug-in corresponds to a lightweight extension, while anchor re-centering corresponds to a full offline update of the R\textsuperscript{2}M decomposition without end-to-end retraining.

\noindent\textbf{Results.}
\Cref{fig:incremental_emerging} shows different behaviors for nearby and novel emerging families.
For UniFace, the base system already achieves near-perfect performance, indicating that the original FF residual geometry covers this family well.
Consequently, adding a plug-in branch or re-centering the anchor brings little additional benefit.
For MidJourney, however, the base system is much weaker, showing that the new family is not well represented by the original anchor-centered residual geometry.
Fast plug-in is insufficient because it adds a new branch while keeping the old anchor and residual subspace fixed.
In contrast, anchor re-centering substantially improves MidJourney while preserving strong performance on the original FF families.

These results illustrate how R\textsuperscript{2}M can support incremental deployment.
Nearby fake families can often be handled by the existing merged detector, while more novel families may require re-centering the anchor and rebuilding the residual decomposition.
Importantly, both update modes operate at the model-merging level and do not require end-to-end retraining of the unified detector.

\end{document}